\documentclass{article}

    \usepackage[final,nonatbib]{neurips_2024}

\usepackage[utf8]{inputenc} 
\usepackage[T1]{fontenc}    
\usepackage[pagebackref=true,breaklinks=true,colorlinks,bookmarks=false]{hyperref}
\usepackage{url}            
\usepackage{booktabs}       
\usepackage{amsfonts}       
\usepackage{nicefrac}       
\usepackage{microtype}      

\usepackage{multirow}
\usepackage{tikz}
\usepackage{colortbl}
\usepackage{algorithm}
\usepackage{algpseudocode}
\usepackage{pifont}

\usepackage{wrapfig}
\usepackage{makecell} 
\usepackage{caption} 

\usepackage{sidecap}
\usepackage{graphicx}
\usepackage[export]{adjustbox}

\usepackage{amsmath}
\usepackage{subcaption}

\usepackage[utf8]{inputenc}
\usepackage{tcolorbox}
\usepackage{geometry}
\usepackage{listings} 
\usepackage{mdframed}

\newcommand{\assembler}{Timeline Assembler}
\newcommand{\vst}{VIST-A}
\newcommand{\vista}{VID-A}
\newcommand{\cmark}{\ding{51}}%
\newcommand{\xmark}{\ding{55}}%

\newcommand{\eg}{\textit{e.g.,}\ }

\newcommand{\etal}{\textit{et al.}}

\usepackage{authblk}

\title{Generative Timelines \\ for Instructed Visual Assembly}

\author[1,2]{\textbf{Alejandro Pardo}}
\author[2]{\textbf{Jui-Hsien Wang}}
\author[1]{\textbf{Bernard Ghanem}}
\author[2,3]{\textbf{Josef Sivic}}
\author[2]{\textbf{Bryan Russell}}
\author[2]{\textbf{Fabian Caba Heilbron}}

\affil[1]{AI Initiative, KAUST}
\affil[2]{Adobe Research}
\affil[3]{CIIRC CTU\thanks{Czech Institute of Informatics, Robotics and Cybernetics at the Czech Technical University in Prague.}}
\affil[ ]{\url{https://sites.google.com/kaust.edu.sa/timeline-assembler}}

\title{Generative Timelines \\for Instructed Visual Assembly}

\begin{document}

\maketitle

\footnotetext[1]{Work partially done during an internship at Adobe.}
\begin{figure}[ht]
    \centering
    \vspace{-.5cm}
    \includegraphics[width=\textwidth]{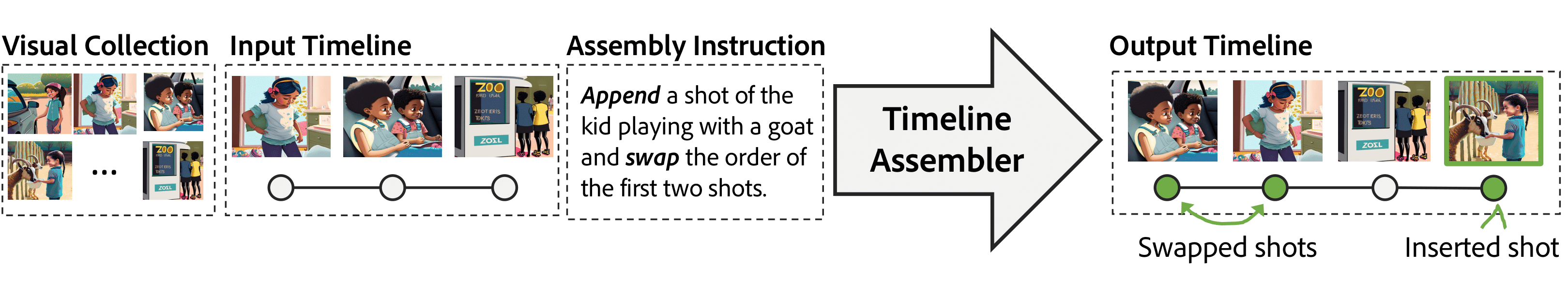}
    \vspace{-.5cm}
    \caption{\textbf{Instructed Visual Assembly.} Given a visual collection, an input timeline, and an assembly instruction, our model (called the \assembler{}) performs the instructed assembly task and generates an output timeline with the desired edits. The collection comprises various media elements, such as video clips or images. The timeline is a sequential arrangement of these elements. }
    \label{fig:pull_figure}
\end{figure}
\begin{abstract}

The objective of this work is to manipulate visual timelines (\eg a video) through natural language instructions, making complex timeline editing tasks accessible to non-expert or potentially even disabled users. We call this task {\em Instructed visual assembly}. This task is challenging as it requires (i) identifying relevant visual content in the input timeline as well as retrieving relevant visual content in a given input (video) collection, (ii) understanding the input natural language instruction, and (iii) performing the desired edits of the input visual timeline to produce an output timeline. To address these challenges, we propose the \assembler{}, a generative model trained to perform instructed visual assembly tasks. The contributions of this work are three-fold. First, we develop a large multimodal language model, which is designed to process visual content, compactly represent timelines and accurately interpret timeline editing instructions. Second, we introduce a novel method for automatically generating datasets for visual assembly tasks, enabling efficient training of our model without the need for human-labeled data. Third, we validate our approach by creating two novel datasets for image and video assembly, demonstrating that the \assembler{} substantially outperforms established baseline models, including the recent GPT-4o, in accurately executing complex assembly instructions across various real-world inspired scenarios.
\end{abstract}
    
\section{Introduction}
\label{sec:intro}

Imagine returning from a zoo trip and finding a visual timeline, \eg in the form of a short video, of highlights automatically generated by your device \cite{nytimes-googlephotos,fastcompany2024}. While reviewing the timeline, you notice some shots are misplaced and a memorable moment with that kid petting a goat is missing. Traditionally, making such modifications would require finding the additional shot in your video collection as well as potential cumbersome interactions with video editing tools. This may be even harder for users with small-screen devices or users with disabilities for whom it is hard to interact with traditional interfaces. A more intuitive approach, illustrated in Figure~\ref{fig:pull_figure}, involves using natural language to direct edits. The users simply state their desired changes, much like setting a calendar event by voice. Such a system uses the collection of visual assets, the current timeline, and the provided instruction to return the refined timeline with the requested edits.

We define this capability as instructed visual assembly, a process that involves the automated editing of a visual timeline in response to user-provided natural language instructions. To effectively automate this task, the system requires a comprehensive understanding of three key elements: the instruction itself, the existing visual timeline, and the collection of visual assets. For instance in Figure~\ref{fig:pull_figure}, if a user instructs the system to \textit{"swap the first two clips and add the shot of the kid with the goat,"} the system must first understand the multimodal context by interpreting the wording of the instruction and relating it to the visual data in the timeline and the collection. The system must then identify the specific elements to be edited, such as the initial two shots, and retrieve the additional clip of the child with the goat from the collection. This process involves deep multimodal understanding to determine not only what the user intends but also how these intentions translate into direct manipulation of visual content. Despite these challenges, instructed visual assembly can enable intuitive video creation interfaces for novices, and users with disabilities, mitigating the complexities associated with mastering traditional video editing tools and interfaces, as well as easing the management of visual content. Moreover, it can ease the editing of videos on devices with small screens where simple operations like drag and drop become cumbersome.

This is the first work that explicitly addresses the task of instructed visual assembly. While this task shares similarities with language-based video editing \cite{fu2022m3l, qin2023instructvid2vid}, which consists of editing the pixels of a source video based on language instructions, instructed visual assembly distinctively focuses on executing instructions to arrange visual elements in a timeline. There are other related tasks in the video assembly space, such as automatic shot transitioning \cite{pei2023automatch,pardo2021learning} and B-roll recommendation \cite{huber2019bscript,xiong2022transcript,truong2016quickcut}. However, these tasks primarily provide creative guidance, diverging from the instructed visual assembly goal of composing timelines through natural language instructions.

In this paper, we present a method for teaching a generative model, named \assembler{}, to follow assembly instructions and generate timelines with the appropriate edits. The \assembler{} builds on Large Language Models (LLMs) to leverage their remarkable skills in following instructions \cite{wei2021finetuned, wang2022self} and interpreting multimodal content \cite{liu2023visual, zhu2023minigpt}.  
Specifically, we adapt a multimodal LLM \cite{zhu2023minigpt} to handle the nuances of instructed visual assembly. This adaptation presents two major challenges. Firstly, existing multimodal LLMs are designed to process a single image or video. Thus, devising a representation for handling visual collections and timelines using LLMs remains an open challenge.
Secondly, as shown by our experiments, existing multimodal LLMs struggle to understand and execute visual assembly instructions, hinting that further tuning for the assembly task is required. 
However, we face the challenge of doing so without human-labeled data, which is hard to acquire at scale for such specialized tasks.

Our work makes the following contributions to address the challenges above:

(1) We design a multimodal LLM architecture to process visual collections, encode timelines, and ingest natural language instructions to generate edited output timelines. Our architecture represents each image or video clip in the collection with a unique identifier and a visual representation compatible with the LLM's input space \cite{liu2023visual}. Secondly, we encode the timeline using the unique identifiers of each arranged visual element. Finally, we map the LLM's output tokens directly back to a timeline, associating each token to its relevant visual element. This design not only ensures a compact representation of the visual collection and the timeline, but also facilitates a straightforward reconstruction of timelines from the model's output tokens.

(2) We propose a new approach to automatically generate a paired dataset (input/output) for instructed visual assembly. Our method programmatically creates a collection, input/output timelines, and assembly instructions from an input visual sequence and task candidates. Using this approach, we generate data to learn the projection layer and Low-Rank Adapters (LoRA) \cite{hu2021lora} for effective LLM performance on assembly tasks. Our training approach is not only human-labels-free but also efficient and enables a wide range of timeline editing tasks.

(3) We construct two datasets, one for image sequence assembly (of still images) and one for video assembly, to evaluate instructed visual assembly. On both datasets, our method substantially outperforms baseline approaches, including powerful LLMs such as GPT3.5, highlighting the benefits of our approach. Moreover, we show that our model can match or even exceed the performance of specialized (single-task) assembly models, perform equally well regardless of the length of the timelines, and execute multiple complex instructions at a time.

\section{Related Work}
\label{sec:related}

\noindent\textbf{Multimodal LLMs as Multi-task Interfaces.}
Our work shares a similar spirit with recent trends that use LLM capabilities to perform multiple tasks for different applications \cite{hao2022language}. Several methods have been proposed for image and language tasks~\cite{zhu2023minigpt,alayrac2022flamingo,liu2023visual,li2023otter,liu2023llava,kwon2023image}, video understanding~\cite{maaz2023video,li2023videochat,chen2023videollm,song2023moviechat,liu2023one}, multimodal understanding~\cite{yin2023lamm,zhang2023video}, recommendation systems \cite{geng2023vip5}, and robotics and motion planning~\cite{jiang2023vima,lin2023text2motion,brohan2023rt,driess2023palme,huang2023instruct2act}. Our Timeline Assembler distinguishes itself by specializing in the domain of visual timelines, interpreting natural language to modify sequences of visual data. This application extends the use of LLMs beyond physical tasks to the handling of multimedia content.

\noindent\textbf{Video Assembly.}
Several computer vision techniques have been proposed for automated video editing tasks \cite{rao2022temporal,pei2023automatch,shen2022autotransition,arev2014automatic,leake2017computational,pardo2021learning,chen2023match,rao2020unified}. 
Likewise, multiple works have proposed to tackle video assembly. Li et al.~\cite{lirepresentation} propose multi-shot vlog assembly, drawing parallels to our approach where an initial sequence guides subsequent shot selections from a candidate pool. Furthermore, other works~\cite{wang2019write, wang2020story,xiong2022transcript,lu2023show} have developed tools that assemble videos from input queries, leveraging multiple components for an intuitive video creation experience. While these works emphasize on providing creative guidance, our model introduces a user-guided approach to video assembly, enabling a novel interface to manipulate and arrange timelines with language instructions.

\noindent\textbf{Instruction Fine-tuning and Aligning with User Intent.}
Our paper builds on methods that adapt language models to closely following human instructions \cite{radford2019language,brown2020language}. Wei \etal \cite{wei2021finetuned} first introduced instruction-tuning, enhancing the usability and multi-tasking of LLMs. \cite{ouyang2022training} expanded this by integrating Reinforcement Learning from Human preferences (RLHF). \cite{wang2022self} demonstrated that GPT3-generated instructions could achieve results similar to InstructGPT through self-instruct fine-tuning. 
Our work uses similar ideas and tailors instruction fine-tuning for visual assembly tasks. We do so by automatically gathering novel visual assembly instruction data that serves as training data to our LoRA \cite{hu2021lora} fine-tuned multimodal LLM.

\section{\assembler{}}
\label{sec:method}

\begin{figure*}[t]
    \centering
    \includegraphics[width=0.95\linewidth]{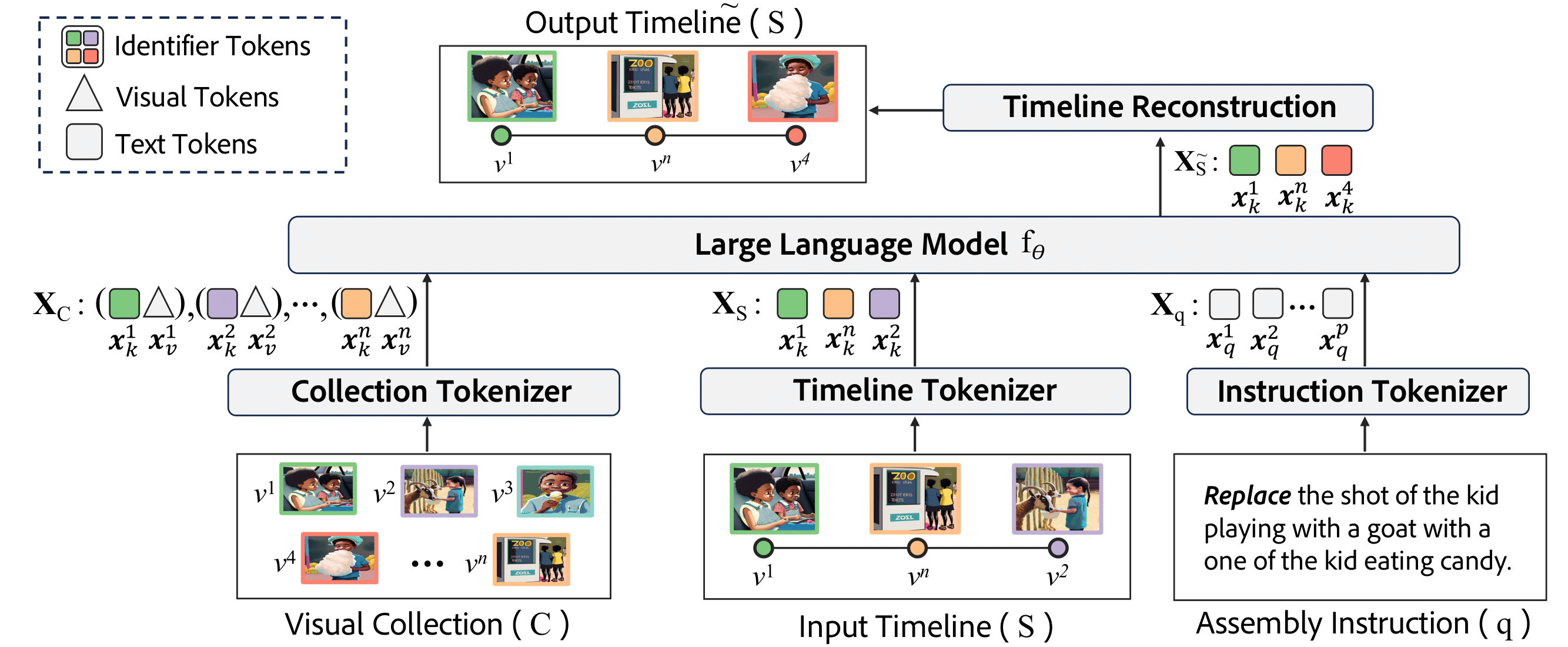}
    \captionof{figure}{\textbf{\assembler{} Architecture.} We design a multimodal architecture to execute visual assembly instructions to generate visual timelines. Our model takes as inputs: a collection of images/videos $C$, a timeline $S$, and an assembly instruction $q$. Each image/video in the collection is represented with a unique identifier token $\mathbf{x}^{i}_{k}$ (color-coded) and a visual token $\mathbf{x}^{i}_{v}$, forming the sequence $\mathbf{X}_C$. The input timeline is represented with the sequence of tokens $\mathbf{X}_S$, which comprises the list of identifier tokens of the images/videos in the timeline. The assembly instruction is tokenized into $\mathbf{X}_q$. Given the input tokens, the task of the Large Language Model is to generate output timeline tokens $\mathbf{X}_{\tilde{S}}$, which are reconstructed into the output timeline $\tilde{S}$.}
    \label{fig:pipeline}
    \vspace{-15pt}
\end{figure*}

\subsection{\assembler{} Architecture}
Our goal is to design the \assembler{} to integrate visual and textual data to generate output timelines that precisely align with the given assembly instructions. The \assembler{} generates an edited timeline $\tilde{S}$ from an assembly instruction $q$, an initial timeline $S$, and a collection of video clips $C$. With instructions often phrased in multiple ways, ensuring consistent output results poses a considerable challenge.  To tackle these challenge, the \assembler{} leverages an instruction-following Large Language Model (LLM) that reasons over text and visual tokens. Our overall architecture, illustrated in Figure \ref{fig:pipeline}, comprises a set of tokenizers that convert multimodal inputs into a set of tokens $\mathbf{X}$. These tokens are then passed to an LLM, which returns an output sequence of tokens $\mathbf{X_{\tilde{S}}}$ that represent a timeline in the LLM space. Finally, we reconstruct the output timeline $\tilde{S}$ from $\mathbf{X_{\tilde{S}}}$. We describe each of these components in detail next. 

\noindent\textbf{Collection Tokenizer.}
Our goal is to represent a collection of visual elements as an array of unique identifier tokens and visual tokens to enable an LLM to process the visual content in the collection. Each identifier token serves to distinctly identify the visual elements within the collection, while the visual tokens encapsulate the visual information. With this design, the LLM can reference the visual representations associated with each element in the collection.

Let $C = \{v^1, v^2, \ldots, v^n\}$ denote the visual collection, where $v^i$ represents a visual element composed of an image or video clip. The collection tokenizer has two key components:  a mapping function $\mathcal{H}$ that generates a unique identifier token $\mathbf{x}^{i}_{k}$ for each visual element $v^i$; and a visual encoder that embeds every visual input $v^i$ to produce a visual token  $\mathbf{x}^{i}_{v}$. The mapping function $\mathcal{H}$ obtains a unique text token $\mathbf{x}^{i}_{k}$ by assigning a token from a set of previously tokenized integers. In practice, this mapping function operates as a look-up table that helps to assign (and find) the unique identifier token of a given visual element. The visual encoder ingests a visual element $v^i$ and outputs a visual token $\mathbf{x}^{i}_{v}$. In practice, a pretrained visual encoder $g$ ingests and extracts visual representations from the input visual elements, and a projection layer $h_{\gamma}(\cdot)$ maps these visual representations into a visual token that is aligned with the input space of the LLM \cite{merullo2022linearly}. Each visual element $v^i$ is then represented as a tuple $(\mathbf{x}^{i}_{k}, \mathbf{x}^{i}_{v})$ consisting of a unique identifier token and a visual token, respectively, such that the entire array of collection tokens $\mathbf{X}_{C}= [(\mathbf{x}^1_k, \mathbf{x}^1_v), (\mathbf{x}^2_k,\mathbf{x}^2_v), \ldots, (\mathbf{x}^n_k, \mathbf{x}^n_v)]$.

\noindent\textbf{Timeline Tokenizer.}
The goal of the timeline tokenizer is to map a sequence of visual elements within a timeline to their corresponding identifier tokens, allocated during the collection's initial tokenization. By utilizing these pre-existing tokens, the tokenizer avoids the need for re-tokenizing visual elements as they appear in the timeline. Doing so prevents redundant visual tokenization, enabling the LLM to handle in practice more extensive collections and longer timelines. As a result, the visual elements (images or videos) are represented only once to save space. Their corresponding identifier token acts as a reference (or pointer) to the (typically high-dimensional) visual token. 

In detail, let $S = \{v^{S1}, v^{S2}, \ldots, v^{Sl}\}$ denote the input timeline, where $v^{Si}$ denotes the $i$-th visual element in the timeline, and $l$ is the length of the timeline. Using the mapping function $\mathcal{H}$, we can retrieve the identifier token for each visual element $v^{Si}$ and construct the tokenized timeline $\mathbf{X}_{S}=[\mathbf{x}^{S1}_{k}, \mathbf{x}^{S2}_{k}, \ldots,  \mathbf{x}^{Sl}_{k}]$.

\noindent\textbf{Instruction Tokenizer.}
Given an input instruction text prompt $q$, the goal of the tokenizer is to map the sequence of strings/words into a discrete set of $p$ tokens such that: $\mathbf{X}_{q} = [\mathbf{x}^1_q, \mathbf{x}^2_q, \ldots,  \mathbf{x}^p_q],$

where $\mathbf{X}_{q}$ is a sequence of tokens that represents the full text prompt $q$.
In practice, we leverage the byte-level BPE text tokenizer \cite{sennrich2015neural} of our large language model.

Finally, the input to the LLM is the union of token sequences $\mathbf{X}$, defined as $\mathbf{X} = [ \mathbf{X}_C, \mathbf{X}_S, \mathbf{X}_q ]$.

\noindent\textbf{Large Language Model.}
Our goal is to develop a model that can assemble and edit timelines of visual sequences. The model must be able to understand natural language instructions and a multimodal context representing a visual collection and the input timeline. To do so, we leverage multimodal LLMs given their capabilities at managing multiple multimodal tasks \cite{liu2023visual, liu2023llava}, reasoning over long sequences \cite{chen2023videollm, lin2023mm}, and encapsulating knowledge from a plethora of sources \cite{ouyang2022training, touvron2023llama}. 

Therefore, we employ a Large Language Model \( f_{\theta}(\cdot) \), parameterized by \( \theta \), to generate an updated timeline sequence \( \mathbf{X}_{\tilde{S}} \) in response to an input instruction. The model takes a sequence of previously defined multi-modal tokens \( \mathbf{X} \) as input and, at test time, outputs a sequence of identifier tokens representing the updated timeline's visual elements: \( f_\theta(\mathbf{X}) \rightarrow \mathbf{X}_{\tilde{S}} = [\mathbf{x}^{\tilde{S}1}_k, \mathbf{x}^{\tilde{S}2}_k, \ldots, \mathbf{x}^{\tilde{S}Q}_k] \), where \( Q \) is the output timeline length, and \( \mathbf{x}^{\tilde{S}i}_k \) is the identifier token for the \( i \)-th element in the updated timeline.

\noindent\textbf{Timeline Reconstruction.} Given the output tokens $\mathbf{{X}}_{\tilde{S}}$, produced by the LLM $f_{\theta}(.)$, the goal of this step is to reconstruct the output timeline $\tilde{S}$. To do so, we map each output token, which in practice are identifier tokens, to their corresponding visual elements using a reverse mapping operation from $\mathcal{H}$.

\subsection{Constructing Visual Assembly Tasks} 
\label{subsec:dataset}

Our goal is to train the \assembler{} to effectively perform various visual assembly tasks. To facilitate this, we define a suite of atomic operations that act as the foundational elements for constructing and manipulating visual timelines. These operations are:
\begin{itemize}
    \item \textit{Insert ($in$)}: Add an element from the collection to the timeline.
    \item \textit{Remove ($rm$)}: Delete an entry from the timeline.
    \item \textit{Replace ($rp$)}: Substitute one element in the timeline with another from the collection.
    \item \textit{Swap ($sw$)}: Exchange positions of two elements within the timeline.
\end{itemize}

To reference elements, either in the timeline or the collection, we use two types of cues:
\begin{itemize}
    \item \textit{Positional Cue ($p$)}: Refers to an element by its identifier or position in the timeline.
    \item \textit{Semantic Cue ($r$)}: Refers to an element through a language-based description of its visual content.
\end{itemize}

These operations and cues combine to form eight distinct assembly tasks $T={t_c}$, where $t$ represents the operations { $in$, $rm$, $rp$, $sw$ } and $c$ denotes the types of cues { $p$, $r$ }. An illustration of each one of these tasks can be found in Figure~\ref{fig:dataset_creation}.
To generate training data for these tasks, we apply a transformation function $\phi_{t_c}$ to an initial timeline ${S}^i$. This function manipulates the timeline to produce a modified version $\tilde{S}^i$. For example, to train the model for the "remove" operation, we artificially introduce an additional shot into the timeline. An automatically generated caption explicitly instructs the model to identify and remove this newly added shot, thus reverting the timeline to its original form. This direct instruction ensures the model learns the specific task through controlled adjustments. Each transformation is guided by instruction templates $q_t$, filled with the applicable cues $c^i$, to ensure the instructions are clear and relevant to the task at hand. We use this procedure to generate data with multiple lengths, and multiple instructions at a time that combine multiple atomic operations in one instruction, as we will show later in the experiment Section~\ref{sec:experiments}. We explain the data generation procedure in more detail in Section~\ref{subsec:dataset} of the supplementary material.

\begin{SCfigure}[][t]
    \centering
    \sidecaptionvpos{figure}{c} 
    \includegraphics[width=0.55\linewidth]{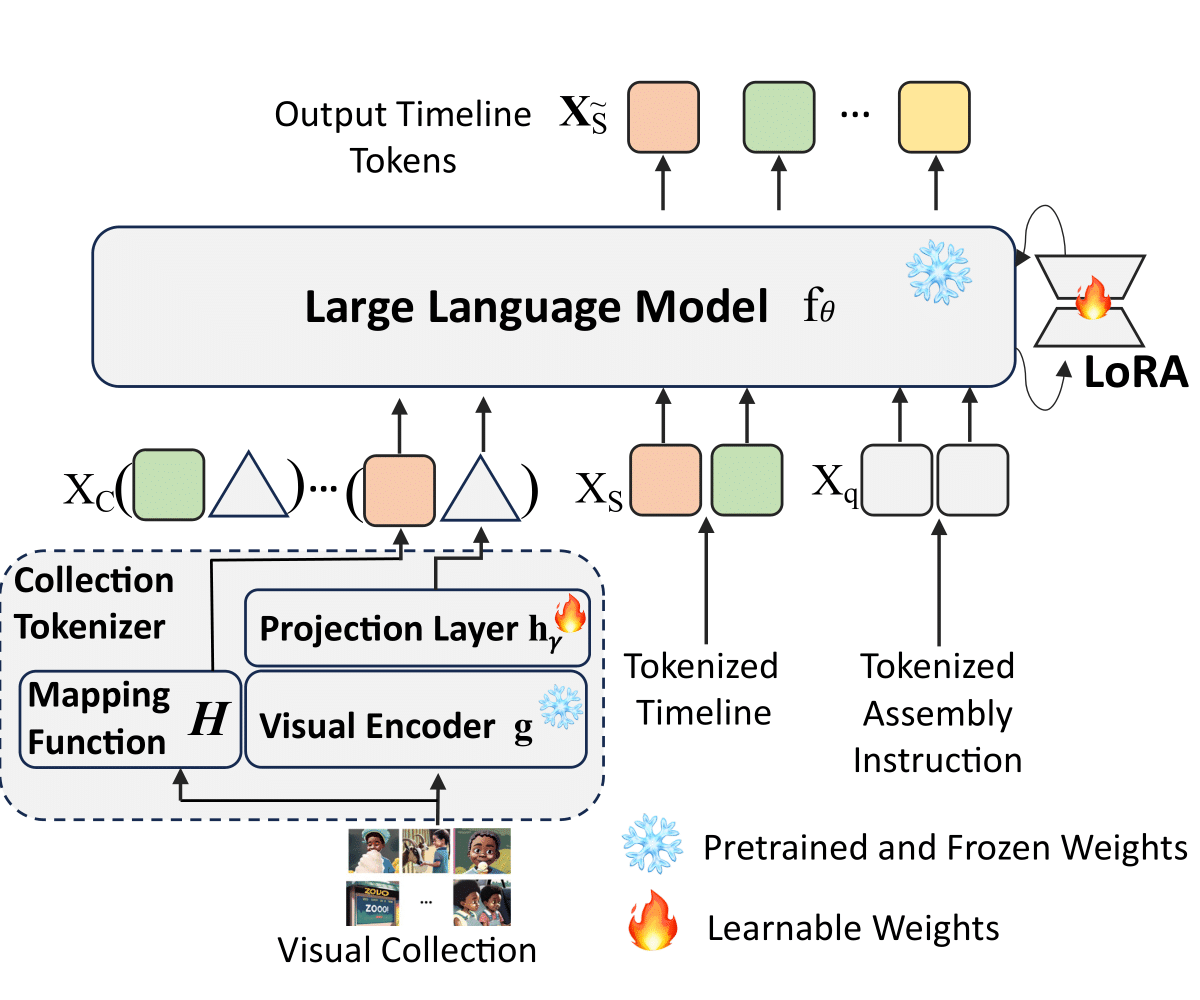}
    \caption{
    \textbf{Training the \assembler{}.}
    The Collection Tokenizer is composed of a frozen visual encoder $g(\cdot)$, a mapping function $\mathcal{H}$ that generates an identifier token for each input visual asset (image/video) in the collection, and a learnable projection layer $h_{\gamma}(\cdot)$ that maps visual embeddings into visual tokens aligned with the LLM. We keep the LLM $f_{\theta}(\cdot)$ mostly frozen except for a lightweight set of learnable LoRA Adapters~\cite{hu2021lora}.
    }
    \label{fig:training}
\end{SCfigure}
\subsection{Training the \assembler{}}

We illustrate our training procedure in Figure \ref{fig:training}. The \assembler{}'s model $\mathcal{M}_{\gamma, \theta}$ has two learnable modules: the projection layer $h_{\gamma}(\cdot)$ and the language model $f_{\theta}(\cdot)$. Since we want to keep the multitask, instruction-following capabilities of the LLM, $f_{\theta}(\cdot)$, we keep its weights mostly frozen except for a lightweight set of learnable Low-Rank Adapters (LoRA)~\cite{hu2021lora}. 

Our goal is to fine-tune the weights of the projection layer and large language model using our proxy assembly tasks gathered in the training dataset $\mathcal{D}$. In practice, we optimize the negative log-likelihood loss (NLL) as cited in Brown et al. (2020) by using the following: $\text{Loss}(\theta, \gamma) = -\sum_{i=1}^{N} \log P(\mathbf{\hat{X}}^i | \mathbf{X}^i; \theta, \gamma)$.
where $N$ is the number of samples and $P(\mathbf{\hat{X}}^i | \mathbf{X}^i; \theta, \gamma)$ is the probability assigned by the model $\mathcal{M_{\gamma, \theta}}$ to the correct output sequence $\mathbf{\hat{X}}^i$ given the input  $\mathbf{X}^i$ formed by the collection tokens $\mathbf{X}_{C}^i$, timeline tokens $\mathbf{X}_{S}^i$, and assembly instruction tokens $\mathbf{X}_{q}^i$ for sample $i$.
\section{Experiments}
\label{sec:experiments}

\noindent\textbf{Implementation Details.}
Our model has four main components as shown in Figure \ref{fig:training}: the visual backbone, the projection layer, the LLM, and the low-rank matrices from LoRA \cite{hu2021lora}. As shown in Figure \ref{fig:training}, the visual backbone and the LLM are kept frozen throughout the training. We only train the projection layer and the LoRA parameters. For the visual encoder, we adopt the same architecture as BLIP-2 \cite{li2023blip} with ViT-g/14 from EVA-CLIP \cite{fang2023eva} and a Q-former that outputs 32 tokens per image \cite{li2023blip}. We initialize the projection layer with that from MiniGPT-4 \cite{zhu2023minigpt}. Finally, for the instruction-following LLM, we use Vicuna \cite{zheng2023judging}. For the visual tasks, we use BLIP-2 Flan T5 XL \cite{li2023blip} captions as cues $c$. We provide additional details in Section~\ref{sec:implementation} in the supplementary material.

\noindent\textbf{Evaluation Metrics.}
Our system's performance is assessed using \emph{assembly accuracy}, which determines whether a generated timeline exactly matches the corresponding ground-truth timeline. A true positive is a timeline with all its elements identical to the ground-truth, while a false positive occurs if any element in the predicted timeline differs from the ground-truth. We calculate the \emph{Overall} assembly accuracy by evaluating each instance in the testing datasets and computing the percentage of correctly predicted samples. Additionally, we report the assembly accuracy for each cue type.

\subsection{New Datasets for Instructed Visual Assembly }

We present the details of our Instructed Assembly Datasets. For both cases, the size of the collection is 20 visual assets (images or videos). However, the design is not limited to this number (we show results with different collection sizes in the supplementary material). For the timeline length, we use $l=5$ for Table~\ref{table:Main}, and $2<=l<=19$ for Section~\ref{subsec:capabilities}. Every task implies a single modification to the timeline. We use two data sources $\Tilde{D}$ to create Instruction Visual Assembly Datasets. Both are described below. To ensure instruction-phrasing generalization, we ensure that the training templates do not overlap with the validation and testing ones.

\noindent\textbf{Visual Storytelling Assembly Dataset (\vst{}).}
We use images from \cite{huang2016visual} as source of image sequences to form our new Visual Storytelling Assembly dataset. The original dataset in \cite{huang2016visual}, contains Flickr images linked to each other by annotators, to create visual stories each consisting of 5 images and their captions. We use the procedure explained in Section~\ref{Alg:DatasetCreation} to generate offline samples for each one of the assembly tasks. For testing, we create 80 samples per task for a total of 640 visual storytelling assembly tasks. For training, we create data online.

\noindent\textbf{Video Sequence Assembly Dataset (\vista{}).}
We collect a total of $12,088$ YouTube Shorts \cite{youtube_shorts}. We use these videos to show the capabilities of the \assembler{} for video assembly. We divide each video into shots and take the center frame as the shot representation. We create all possible timelines of a video by sampling every possible sequence of n consecutive shots. \textit{The visual collection is formed by the shots of the video and when needed}. We construct the \emph{Video Sequence Assembly Dataset (\vista)} by creating 80 samples per task for a total of 640 video assembly tasks for testing. We create three validations sets using this procedure, one with timeline length of 5 for Section~\ref{subsec:main}, one with variable timeline lengths for multi-len experiments in Section~\ref{subsec:capabilities}, and one with several instructions and variable timeline lengths for compositional experiments in Section~\ref{subsec:capabilities}. For training, we generate data samples using Algorithm~\ref{Alg:DatasetCreation} online.

\subsection{Instructed Visual Assembly Results}\label{subsec:main}

\begin{table}[t!]
\centering
\footnotesize
\setlength{\tabcolsep}{4pt}
\renewcommand{\arraystretch}{1.2}
\caption{\textbf{Instructed Visual Assembly Results.} We compare the \assembler{} against multiple baseline approaches. We include zero-shot results for powerful open-source and private VLMs. We compare the performance of our Assembler across various model capacities on two novel datasets, VST and VISTA, reporting assembly accuracy for Overall, Positional, and Semantic (cues). * denotes adjustment on the original implementation of the models.}
\label{table:Main}
\begin{tabular}{l|cc|c||cc|c}
\toprule
\multicolumn{1}{c|}{} & \multicolumn{6}{c}{\textbf{Assembly Accuracy(\%)}} \\

& \multicolumn{3}{c||}{\textbf{VIST-A} (Image-based)} & \multicolumn{3}{c}{\textbf{VID-A} (Video-based)} \\
&  \textbf{Positional} & \textbf{Semantic} & \textbf{Overall} & \textbf{Positional} & \textbf{Semantic} & \textbf{Overall} \\
\midrule

\multicolumn{7}{l}{\textit{Zero-shot}}\\
\hline
{MiniGPT-4\cite{zhu2023minigpt} *} & \textcolor{gray}{0.0} & \textcolor{gray}{0.0} & \textcolor{gray}{0.0}  & \textcolor{gray}{0.0} & \textcolor{gray}{0.0} & \textcolor{gray}{0.0}\\
{LLaVA-1.5\cite{liu2023llava} *} & 5.6 & 0.0 & 2.8  & 4.7 & 0.3 & 2.5\\
GPT-4o & 72.8 & 25.0 & 48.9 & 72.8 & 18.8 & 45.8 \\

\rowcolor[gray]{0.9}\multicolumn{7}{l}{\textbf{Ours}} \\
\hline
\assembler{}-7B & 90.1 & 58.1 & 74.1 & 91.8 & 41.8 & 66.8 \\
\assembler{}-13B & \textbf{96.9} & \textbf{66.4} & \textbf{81.6} & \textbf{93.7} & \textbf{47.5} & \textbf{70.6} \\
\bottomrule
\end{tabular}
\end{table}

In Table~\ref{table:Main}, we compare our model with state-of-the-art multimodal models. We observed that both MiniGPT-4 and LLaVA-1.5 lacked the capacity to manage a vast amount of visual data simultaneously. Accordingly, we adjusted these models to handle the entire collection at once; for more details on these adjustments, please refer to the supplementary material~\ref{sec:additional-results}. After modifying both models for our task, neither showed satisfactory performance in these challenging tasks. We discuss the failure cases of these two models further in Section~\ref{sec:additional-results} of the supplementary material. We attribute their inadequate performance primarily to the data and instructions on which they were trained, which significantly differ from the task of instructed visual assembly.

To establish a stronger baseline, we evaluated the recent GPT-4o and found that this model achieved very impressive zero-shot performance on both datasets, recording $48.9\%$ on \vst{} and $45.8\%$ on \vista{}. However, when trained specifically for the task, our \assembler{} clearly emerged as the best alternative for instructed visual assembly, outperforming all models in both positional and semantic tasks. The smaller \assembler{}-7B achieves $66.8\%$ on the \vista{} dataset. When scaled up to 13B parameters, the \assembler{} improves to $70.6\%$. Thus, our strategy proves effective and further enhances the capabilities of LLMs and multimodal LLMs for assembly tasks. For a detailed breakdown of the results for each task and additional analysis on the low performance of the baselines, please refer to Table~\ref{table:main-visual-expanded} in the supplementary material.
Additionally, we implemented additional baselines to perform instructed visual assembly using text-only language models, leveraging BLIP-2~\cite{li2023blip} and replacing every image with its corresponding caption. Detailed explanations and results of these additional experiments are presented in Table \ref{table:main-text-expanded} in the supplementary material.

\subsection{Analysis of the \assembler{}}

\noindent\textbf{Impact of Multi-Task Training (Table \ref{table:multitask}).}
 We compare the \assembler{} against specialized models trained for each one of the assembly tasks. We use a similar training strategy but train 8 individual models, one for each assembly task on the \vst{} dataset. To determine which of the 8 models to use for a given instruction, we propose three approaches:

\begin{itemize}
    \item\textit{\underline{Random Selection}}: selects one of the eight models for a given instruction.
 
    \item\textit{\underline{Oracle Task Classifier}}: employs an Oracle Task classifier to choose the model.

    \item\textit{\underline{GPT-4o Classification}}: uses GPT-4o to classify the instruction and select the appropriate model. This classifier achieves an accuracy of $81.6\%$. Details about the performance of the GPT-4o Task classifier are provided in the supplementary material. 
\end{itemize}

\noindent Table~\ref{table:multitask} shows the remarkable performance of the \assembler{}. It is surpassed  by only $5\%$ by the single-task models in conjunction with the Oracle classifier. This finding highlights the \assembler{}'s strong multi-tasking capabilities.
The \assembler{}, being a single model that effectively handles multiple tasks without the need for a task classifier, proves to be highly practical. These results suggest that incorporating more assembly tasks could further enhance the \assembler{} model's efficacy. \\

\begin{table*}[t!]
    \caption{\textbf{Impact of Multi-Task Training.} We present a comparative analysis of our proposed model's performance under single-task versus multi-task training paradigms. For the single-task case, we use one model per assembly task and a Task Classifier to decide which model to use. The Random and Oracle classifiers serve as lower-bound and upper-bound references for the single-task cases, respectively. We use GPT-4o to classify every instruction into one of the tasks. Our model (featured in the last row) is a single model that understands and performs every task and does not need any task classifier. For each task we report the assembly accuracy (\%) on the \vst{} dataset.}
    \label{table:multitask}
    \centering
    \footnotesize
    \renewcommand{\arraystretch}{1.2}
    \setlength{\tabcolsep}{2pt} 

    \begin{tabular}{@{}l|cc|cccc|cccc|c@{}}
        \toprule
        &\textbf{Multi} & \textbf{Task} & \multicolumn{4}{c|}{\textbf{Positional Cues}} & \multicolumn{4}{c|}{\textbf{Semantic Cues}} & \textbf{Avg.} \\
        &\textbf{Task} & \textbf{Classifier} & Insert & Remove & Replace & Swap & Insert & Remove & Replace & Swap &  \\
        \midrule
        Lower-Bound & \xmark & Random & 10.0 & 15.0 & 17.5 & 11.3 & 20.0 & 17.5 & 15.0 & 16.3 & 15.3 \\
        Upper-Bound & \xmark & Oracle & 99.2 & 98.8 & 100.0 & 98.8 & 71.3 & 70.0 & 69.6 & 26.3 & 79.2 \\
        \hline
        Single-Task Models & \xmark & GPT-4o & 88.1 & \textbf{97.6} & 80.0 & 76.5 & \textbf{67.7} & \textbf{70.0} & 46.1& 12.1 & 67.3 \\
        \assembler{} & \cmark & N/A & \textbf{98.8} & 90.4 & \textbf{85.4} & \textbf{85.8} & 65.8 & 66.7 & \textbf{54.2} & \textbf{45.8} & \textbf{74.1} \\
        \bottomrule
    \end{tabular}
\vspace{-10pt}
\end{table*}

\noindent\textbf{Impact of Training Scale.}

We analyze how data size impacts the \assembler{}'s learning. We limit the percentage of \vst{} and \vista{} data samples available during training. Since we create training data on-the-fly, cutting down the number of sequences available during training means less variability in the data. Therefore, we scale the number of epochs accordingly. Table~\ref{table:datasize} shows three sizes for \vst{} and \vista{} datasets. We observe that the availability of more data for creating assembly tasks on-the-fly leads to better performance.

\noindent\textbf{Cross Dataset Analysis.}
Table~\ref{table:xdataset} presents the results when our model is trained on one dataset and tested on the other. Our model demonstrates generalizability across datasets. When the \assembler{} is trained on visual stories, it can still learn how to perform video sequence assembly and vice versa. Complementary to the observations in Table~\ref{table:datasize}, when merging the two datasets during training, the \assembler{} performs the best across both datasets by gaining a notable $6\%$ on \vst{} and $8\%$ on \vista{}. Thus, the favored approach is to continuously incorporate more data.

\noindent\textbf{Ablations.}
Table~\ref{table:ablation} contrasts our model's performance with and without its key components. 

First, when deactivating LoRA (first row), the \assembler{}'s performance drops by a significant $31\%$. 
Second, initializing the projection layer from~\cite{zhu2023minigpt} but freezing it (second row) instead of finetuning it, substantially lowers the assembly accuracy by $12.5\%$. 
Finally, when the projection layer is trained from scratch (third row), the performance is still competitive, as it only drops by $4\%$.
These results validate that training both, the projection layer and LoRA adapters, is crucial in training \assembler{} to perform instructed visual assembly tasks.

\subsection{\assembler{} Capabilities}\label{subsec:capabilities}
In this section, we evaluate the \assembler{} on more complex tasks inspired by real video editing applications. We aim at evaluating the \assembler{} under two difficult scenarios: \textbf{(i)} The first scenario consists of handling variable input timeline lengths. This scenario challenges the \assembler{} with varying-length sequences and increasingly harder tasks as the length increases. \textbf{(ii)} The second scenario challenges the \assembler{} to deal with the composition of instructions specified in a single query. In this task, the model must understand multiple instructions at once and perform several timeline modifications to successfully resolve the user's input.

\noindent\textbf{Multi-Length Timeline Assembler.}
In the previous sections, we evaluated the \assembler{}'s capabilities with fixed input timeline lengths of 5 assets. However, in real-life applications, dealing with multiple timeline lengths is essential. Therefore, using Algorithm~\ref{Alg:DatasetCreation} with the same data source as \vista{}, we construct a test set for multiple-length assembly tasks, which we call multilen-\vista{}. At training time, we also create data on-the-fly using Algorithm~\ref{Alg:DatasetCreation}. Figure~\ref{fig:length-analysis} reports the assembly accuracy of the assembler trained on multiple lengths against the strongest baseline GPT-4o from Table~\ref{table:Main}. We report the average assembly accuracy across all tasks within different ranges of timeline lengths (detailed per-task results can be found in the supplementary material). We observe a decrease in GPT-4o's performance as the input timeline length increases to 5 and above, indicating that the tasks become more difficult. Our \assembler{} performs consistently across different lengths.

\noindent\textbf{Compositional Timeline Assembler.}
Another desirable feature for the assembler is to handle compositions of assembly instructions. This compositional capability would allow greater flexibility in the assembler's functionality. Therefore, we create a test set of compositional semantic assembly tasks, named Compositional-\vista{}. Since semantic tasks are more challenging (as shown in Table~\ref{table:Main}), we combine them for our test set, presenting a highly challenging yet practical scenario for the assembler. Our test set includes combinations of two semantic tasks in a single instruction. Examples of such instructions can be found in Figures~\ref{fig:qualitative_result_001}~to~\ref{fig:qualitative_result_004}.
In Table~\ref{table:compositional}, we present different training strategies and their impact on assembly accuracy for multilen-\vista{} and Compositional-\vista{}. We note that GPT-4o struggles with multiple tasks simultaneously. Notably, the compositional \assembler{} outperforms GPT-4o and manages to perform compositional operations $36.3\%$ of the time. Interestingly, the best way to train the \assembler{} for compositional tasks is to also incorporate single tasks (``Atomic-Task Training'') during training, as shown in rows 2 (compositional only) vs. 3 (compositional and atomic) of Table~\ref{table:compositional}. The Compositional \assembler{} also performs well on Multi-len \vista{} and \vista{}, we report these results in the supplementary material.
\vspace{-10pt}
\begin{table*}[t]
     \caption{\textbf{Analysis of the \assembler{}}. (a) studies the impact of varying the training data size.
    (b) delves into the model's generalization by testing its performance when trained on one dataset and tested on another; `mean' is the average accuracy across both tests. (c) presents an ablation study, indicating the model's performance with different variations of the model on \vst{}.}
    \label{table:Analysis}
    \centering
    \begin{adjustbox}{center}
    \footnotesize
    \setlength{\tabcolsep}{2.0pt}
    \subfloat[\textbf{Training scale}]{\label{table:datasize}
        \begin{tabular}{ccc}
            \textit{Training scale} & \vst{} & \vista{} \\
            \hline
            $10\%$ & 17.2 & 15.9 \\
            $50\%$ & 71.9 & 61.7\\
            $100\%$ & \textbf{74.1} & \textbf{62.2}\\
            \hline
        \end{tabular}
    }
    \hspace{0.1cm}
    \subfloat[\textbf{Cross dataset analysis}]{\label{table:xdataset}
        \begin{tabular}{lcc|c}
            & \multicolumn{2}{c}{\textit{Testing}} & \\
            \textit{Training} & \vst{} & \vista{} & mean \\ 
            \hline
            VIST-A & 74.1 & 62.2 & 68.2 \\
            VID-A & 71.7 & {66.8} & {69.2} \\
            \textit{all} &\textbf{ 80.0} & \textbf{68.8} & \textbf{74.4} \\
            \hline
        \end{tabular}
    }
    \hspace{0.1cm}
    \subfloat[\textbf{Learnable module ablation}]{\label{table:ablation}
    \begin{tabular}{lc}
        & \vst{}  \\
        \hline
        w/o LoRA & 43.0 \\
        frozen $h_{\gamma}(\cdot)$ init. from ~\cite{zhu2023minigpt} & 61.6  \\
        $h_{\gamma}(\cdot)$ from scratch& 70.5 \\
        \hline
        Ours & \textbf{74.1} \\
        \hline
    \end{tabular}
    
    }
    \vspace{-15pt}
    \end{adjustbox}
    \vspace{-18pt}
   
\end{table*}

\noindent\begin{minipage}{.43\linewidth}
    \centering
    \includegraphics[width=\linewidth]{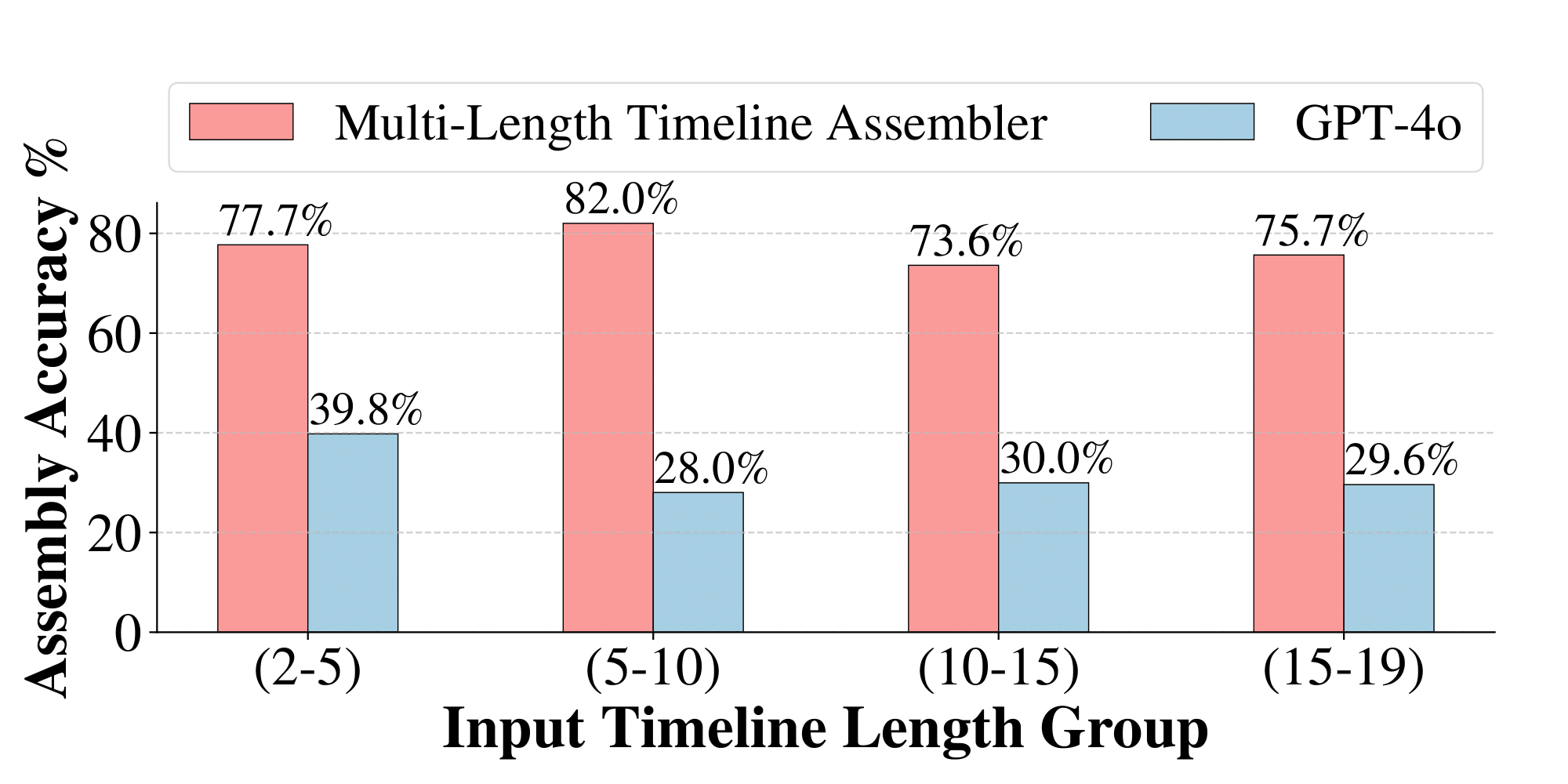}
    \captionof{figure}{\textbf{Multi-Length Timeline Assembler}. Unlike GPT-4o, which has decreasing performance after timeline lengths larger than 5, our \assembler{} is consistent across different lengths.}
    \label{fig:length-analysis}
    \vspace{-10pt}
\end{minipage}%
\hspace{5mm}
\begin{minipage}{.52\linewidth}
    \centering
    \footnotesize
    \setlength{\tabcolsep}{2pt}
    \captionof{table}{\textbf{Compositional Timeline Assembler.} We show results on compositional tasks. Our Compositional \assembler{} outperforms GPT-3.5. Additionally, we show the benefit of including single-tasks (atomic) during training.}
    \label{table:compositional}
    \begin{tabular}{l|cc|c}
    \toprule
    &  &  & Semantic\\
    & \multicolumn{2}{c|}{Training}& Comp. \\
    Model & Atomic & Comp. & \vista{} (\%) \\
    \midrule
    GPT-4o & \xmark & \xmark  & 3.8  \\
    \assembler{} & \xmark & \cmark & 26.3 \\
    \assembler{} & \cmark & \cmark  & \textbf{36.3} \\
    \bottomrule
    \end{tabular}
    \vspace{-15pt}
\end{minipage}

\section{Conclusion}

\vspace{-5pt}
We introduce the \assembler{}, the first generative model for instructed visual assembly. 
We train our model using automatically generated assembly tasks. 
We validate our approach on two newly built instruction visual assembly datasets and show that the \assembler{} follows assembly instructions more accurately than competing strong multi-modal large language models. 
Looking ahead, we believe the \assembler{} is a step towards generative models capable of complex reasoning (such as timeline editing or long-form story telling) over large collections of multi-modal assets. We include the limitations of our paper in~\ref{section:limitations}.

\paragraph{Acknowledgments}
The research reported in this publication was partially supported by funding from King Abdullah University of Science and Technology (KAUST) - Center of Excellence for Generative AI, under award number 5940.
\clearpage
{\small
\bibliography{main}
\bibliographystyle{ieee}
}

\clearpage
\appendix
\section*{\centering Supplementary Material}
\setcounter{section}{0}

In the following document, we present the limitations of our work in Section~\ref{section:limitations}. Additional details on how to construct visual assembly tasks in Section~\ref{subsec:dataset}. In Section~\ref{sec:additional-results}, we present additional results as well as detail explanations of the results presented in the main paper. In Sections~\ref{sec:tokenizers} and~\ref{sec:implementation} present details on the tokenizers, and implementation details of the models, respectively. Finally, Sections~\ref{sec:failure}, and~\ref{sec:qual-supp} present qualitative failure cases and qualitative success cases of the \assembler{}.
 
\renewcommand{\thefigure}{A\arabic{figure}}
\renewcommand{\thetable}{A\arabic{table}}
\section{Limitations}\label{section:limitations}

Our paper's main limitation is efficiency. Generating timelines autoregressive is costly in high-capacity GPUs, let alone low-power devices. However, our work can directly benefit from the efforts of the community to make Language models run faster and more efficiently. Additionally, our model is only limited to the set of operations we have described. Further directions could work on incorporating another type of operations of the timeline like pixel editing or even trimming shots.

\section{Constructing Visual Assembly Tasks} 
\label{sec:dataset}

\begin{wrapfigure}{r}{0.5\textwidth}
\begin{minipage}{0.5\textwidth} 
\begin{algorithm}[H] 
\caption{Creating Visual Assembly Task Datasets}\label{Alg:DatasetCreation}
\begin{algorithmic}[1]
\small
\Require Source Data $\tilde{\mathcal{D}}$
\Require Set of tasks $T$
\Require Dataset Size $N$
\Ensure Dataset $\mathcal{D}$
\State Initialize $\mathcal{D} = \{\}$
\For{$i$ in range($N$)}
    \State Initialize Collection $C^i = \{\}$
    \State choose sequence length $l$
    \State Sample sequence $S^i$ of size $l$ from $\tilde{\mathcal{D}}$
    \State Sample task $T^i$, with operation $t^i$, cue $c^i$, instruction template $q^i_{t}$ and transformation $\phi^i_{t_c}$
    \State $\tilde{S}^i = \phi^i_{t_c}({S^i})$
    \State Fill up instruction template $q^i_{t}$ using cue $c^i$
    \State Fill up Collection $C^i$ using $\tilde{\mathcal{D}}$ and $S$
    \State $\mathcal{D} = \mathcal{D} \cup \{({S^i}, \tilde{S}^i, q^i, C^i)\}$
\EndFor
\State \Return $\mathcal{D}$
\end{algorithmic}
\end{algorithm}
\end{minipage}
\end{wrapfigure}

\begin{figure*}[h!]
    \centering
    \includegraphics[width=0.95\linewidth]{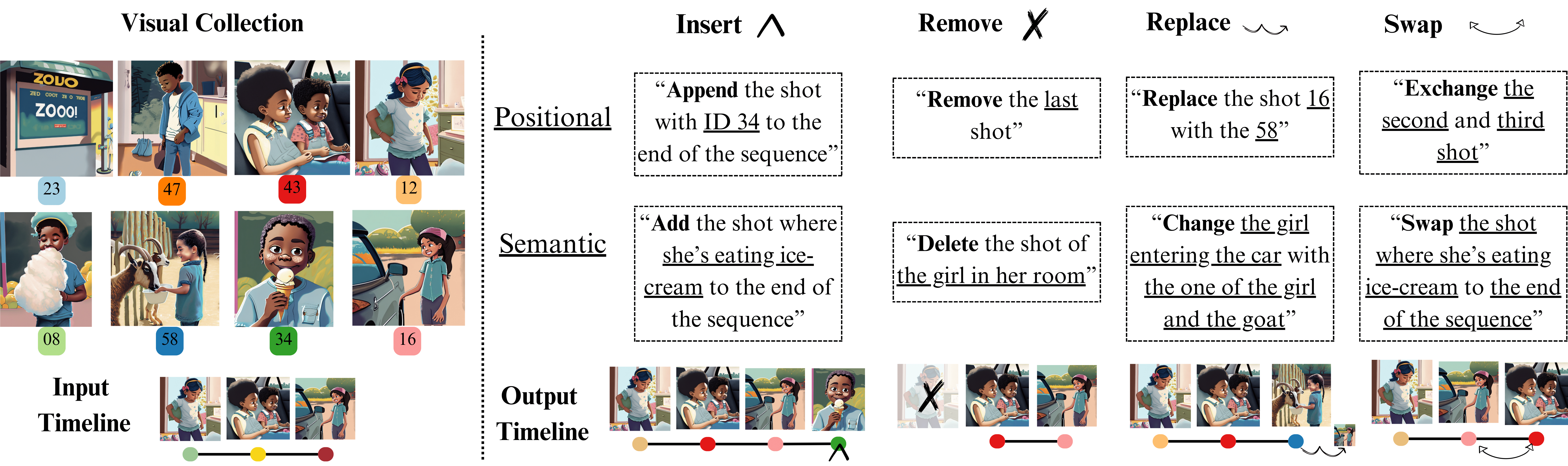}
    \captionof{figure}{\textbf{Visual Assembly Tasks.} The Visual Collection (left) showcases individual visual assets labeled with their corresponding identifiers. The Input Timeline (bottom left) indicates the initial state of the timeline, while the Output Timeline (right) displays the edited timeline after applying the respective editing instructions. Each instructions is described both positionally and semantically, highlighting the system's ability to interpret and execute edits based on shot identification and semantic information. The bold font indicates the assembly operation, while the underlined font indicates the cue (shot ID or a shot description).}
    \label{fig:dataset_creation}
    \vspace{-15pt}
\end{figure*}

We illustrate our assembly tasks in Figure \ref{fig:dataset_creation}. We consider 8 types of basic operations as defined in Section~\ref{subsec:dataset} of the main manuscript. Each one of the tasks is associated with a corresponding transformation function $\phi_{t_c}$ and a set of instruction templates $q_t$. 
Such transformation function $\phi_{t_c}$ is a timeline operation that we use to get a pair of input and target timelines. For instance, the remove function $\phi_{rm_c}$ removes one element from the input timeline ${S^i}$ to get the target timeline $\tilde{S}^i$. On the other hand, the set of instruction templates $q_t$ are templates that are filled up with the given cues $c^i$. For instance in Figure~\ref{fig:dataset_creation}, one of the removing instructions is ``\texttt{Remove the \textit{last} shot}''. It was created using the instruction template $q_{rm}$ ``\texttt{Remove the \{ \} shot}'' and the cue  ``\texttt{last}''. 
Using all the elements previously introduced, we explain how to create an instructed assembly dataset $\mathcal{D}$ next.

\noindent\textbf{Creating Visual Assembly Task Datasets} (Algorithm \ref{Alg:DatasetCreation}). Starting from an existing Data Source $\tilde{\mathcal{D}}$ comprised of visual sequences, a predefined set of assembly tasks $T$, with corresponding transformations $\phi_{t_c}$, and a predefined template instruction $q_t$, we sample a timeline $S^i$ of length $l^i$ from $\tilde{\mathcal{D}}$, an assembly operation $t^i$, a cue $c^i$, and an instruction template $q^{i}_{t}$ corresponding to the assembly task. We then obtain the output timeline $S^i$ using transformation $\phi^i_{t_c}$.
Then, we populate the template $q^{i}_{t}$ using the cue $c^i$. Finally, we fill in the collection $C$ with the $l^i$ visual elements in $\tilde{S}^i$ and $|C|-l^i$ elements from $\tilde{\mathcal{D}}$. We repeat this process $N$ times to obtain a visual assembly dataset $\mathcal{D}=\{ ({S^1}, \tilde{S}^1, q^1, C^1), \dots, ({S^N}, \dots\}$.\\
\section{Additional Results}
\label{sec:additional-results}
\noindent\textbf{Performance Comparison Using BLIP-2 and Text-only Models.} As mentioned before, we perform instructed visual assembly using text-only language models, by replacing each image in the collection with their corresponding BLIP2~\cite{li2023blip} caption. By doing this, the text-only language models can now perform visual assembly tasks, seamlessly. Table~\ref{table:main-text-expanded} presents the results of the text-only models. 

\begin{table*}[ht]
    \centering
    \setlength{\tabcolsep}{5pt}
    \renewcommand{\arraystretch}{1.2}
    \begin{subtable}{\linewidth}
        \centering
        \begin{tabular}{lcccc|cccc|c}
        \cline{2-10}
        \multicolumn{1}{c}{} & \multicolumn{9}{c}{\textbf{Assembly Accuracy (\%)}} \\
        & \multicolumn{4}{c}{\textbf{Positional Cues}} & \multicolumn{4}{c}{\textbf{Semantic Cues}} &  \\
         & Ins. & Rem. & Repl. & Swap & Ins. & Rem. & Repl. & Swap & \textbf{Avg.} \\
        \hline
        \multicolumn{10}{l}{\textit{Zero-shot}} \\
        \hline
        Vicuna-7B~\cite{zheng2023judging} & 18.8 & 7.5 & 48.8 & 7.5 & 2.5 & 6.3 & 13.8 & 5.0 & 13.8 \\
        Vicuna-13B~\cite{zheng2023judging} & 25.0 & 12.5 & 38.8 & 7.5 & 8.8 & 16.3 & 17.5 & 3.8 & 16.3 \\
        GPT-4o & 60.0&	96.3&	63.8&	71.3&	\textbf{50.0}&	\textbf{97.5}&	\textbf{72.5}& \textbf{52.5}& \textbf{70.5} \\
        \hline
        \multicolumn{10}{l}{\textit{Fine-tuned Text-Only}} \\
        \hline
        Vicuna-7B~\cite{zheng2023judging} & 17.9 & 99.6 & \textbf{94.2} & \textbf{98.3} & 0.0 & 20.8 & 4.6 & 11.7 & 43.4 \\
        Vicuna-13B~\cite{zheng2023judging} & \textbf{82.1} & \textbf{100.0} & 80.0 & 97.5 & 4.6 & 20.4 & 6.3 & 5.8 & 49.6 \\
        \hline
        \end{tabular}
        \caption{\vst{}}
        
    \end{subtable}%
    \\
    \begin{subtable}{\linewidth}
        \centering
        \begin{tabular}{lcccc|cccc|c}
        \cline{2-10}
         \multicolumn{1}{c}{} & \multicolumn{9}{c}{\textbf{Assembly Accuracy (\%)}} \\
        & \multicolumn{4}{c}{\textbf{Positional Cues}} & \multicolumn{4}{c}{\textbf{Semantic Cues}} &  \\
         & Ins. & Rem. & Repl. & Swap & Ins. & Rem. & Repl. & Swap & \textbf{Avg.} \\
        \hline
        \multicolumn{10}{l}{\textit{Zero-shot}} \\
        \hline
        Vicuna-7B~\cite{zheng2023judging} & 23.8 & 10.0 & 7.5 & 11.3 & 2.5 & 10.0 & 1.3 & 3.8 & 8.8 \\
        Vicuna-13B~\cite{zheng2023judging} & 21.3 & 11.3 & 7.5 & 10.0 & 10.0 & 12.5 & 7.5 & 6.3 & 10.8 \\
        GPT-4o & 72.5&	93.8&	60.0&	65.0&	\textbf{47.5}&	\textbf{91.3}&	\textbf{71.3}&	\textbf{58.8}&\textbf{70.0} \\
        \hline
        \multicolumn{10}{l}{\textit{Fine-tuned Text-Only}} \\
        \hline
        Vicuna-7B~\cite{zheng2023judging} & 56.3 & \textbf{100.0} & \textbf{97.1} & 91.3 & 2.5 & 24.6 & 6.7 & 8.8 & 48.4 \\
        Vicuna-13B~\cite{zheng2023judging} & \textbf{85.0} & \textbf{100.0} & 91.3 & \textbf{98.3} & 7.5 & 22.5 & 5.0 & 10.4 & 52.5 \\
        \hline
        \end{tabular}
        \caption{\vista{}}
        
    \end{subtable}%
    \caption{\textbf{Performance Comparison Using BLIP-2 \cite{li2023blip} and Text-only Models.} Similar to what we saw in the main manuscript, GPT-4o surpasses the other baselines by a good margin. However, when training with out method, the vicuna models become our text-only \assembler{}s surpassing all the other alternatives. }
    \label{table:main-text-expanded}
\end{table*}

\paragraph{\textbf{Assembly accuracy per task for Table 1. Additional details on MiniGPT4, LLava, and GPT-4o}.} In Table~\ref{table:main-visual-expanded} we showcase an expanded version of Table~1 of the main paper where we show the performance breakdown per task. 
\begin{table*}[ht]
    \centering
    \setlength{\tabcolsep}{2.5pt}
    \renewcommand{\arraystretch}{1.2}
    \begin{subtable}{\linewidth}
        \centering
        \begin{tabular}{lcccc|cccc|c}
        \cline{2-10}
        \multicolumn{1}{c}{} & \multicolumn{9}{c}{\textbf{Assembly Accuracy (\%)}} \\
        & \multicolumn{4}{c}{\textbf{Positional Cues}} & \multicolumn{4}{c}{\textbf{Semantic Cues}} &  \\
         & Ins. & Rem. & Repl. & Swap & Ins. & Rem. & Repl. & Swap & \textbf{Avg.} \\
        \hline
        \multicolumn{10}{l}{\textit{Zero-shot}} \\
        \hline
        MiniGPT-4~\cite{zhu2023minigpt} & 0.0 & 0.0 & 0.0 & 0.0 & 0.0 & 0.0 & 0.0 & 0.0 & 0.0 \\
        Llava-1.5~\cite{liu2023llava} & 12.5&	7.5&	2.5&	0.0&	0.0&	0.0&	0.9&	0.0& 2.9 \\
        GPT-4o & 60.0&	96.3&	63.8&	71.3&	11.3&	30.0&	45.0&	13.8& 48.9 \\
        \hline
        \rowcolor[gray]{0.9}\multicolumn{10}{l}{\textbf{Ours}} \\
        \hline
        \assembler{}-7B & \textbf{98.8} & 90.4 & 85.4 & 85.8 & 65.8 & 66.7 & 54.2 & 45.8 & 74.1 \\
        \assembler{}-13B & 89.2 & 99.2 & \textbf{99.2} & \textbf{100.0} & 54.6 & \textbf{76.7} & \textbf{73.8} & \textbf{60.4} & \textbf{81.6} \\
        \hline
        \end{tabular}
        \caption{\vst{}}
        
    \end{subtable}%
    \\
    \begin{subtable}{\linewidth}
        \centering
        \begin{tabular}{lcccc|cccc|c}
        \cline{2-10}
         \multicolumn{1}{c}{} & \multicolumn{9}{c}{\textbf{Assembly Accuracy (\%)}} \\
        & \multicolumn{4}{c}{\textbf{Positional Cues}} & \multicolumn{4}{c}{\textbf{Semantic Cues}} &  \\
         & Ins. & Rem. & Repl. & Swap & Ins. & Rem. & Repl. & Swap & \textbf{Avg.} \\
        \hline
        \multicolumn{10}{l}{\textit{Zero-shot}} \\
        \hline
        MiniGPT-4~\cite{zhu2023minigpt} & 0.0 & 0.0 & 0.0 & 0.0 & 0.0 & 0.0 & 0.0 & 0.0 & 0.0 \\
        Llava-1.5~\cite{liu2023llava} & 8.8&	7.5&	2.5&	0.0&	0.0&	0.0&	1.3&	0.0& 2.5 \\
        GPT-4o & 72.5&	93.8&	60.0&	65.0&	10.0&	21.3&	27.5&	16.3& 45.8 \\
        \hline
        \rowcolor[gray]{0.9}\multicolumn{10}{l}{\textbf{Ours}} \\
        \hline
        \assembler{}-7B & 75.4 & 98.8 & 95.8 & 97.1 & 42.5 & 55.0 & 41.7 & 27.9 & 66.8 \\
        \assembler{}-13B & \textbf{89.6} & \textbf{100.0} & 96.3 & \textbf{99.6} & \textbf{58.3} & \textbf{67.9} & \textbf{64.6} & \textbf{58.3} & \textbf{79.3} \\
        \hline
        \end{tabular}
        \caption{\vista{}}
        
    \end{subtable}%
    \caption{\textbf{Performance Breakdown Per Task (Table~1 Main Paper Supplemental)} }
    \label{table:main-visual-expanded}
\end{table*}

\paragraph{\textbf{Additional details on MiniGPT-4, LLava, and GPT-4o}} As indicated in Tables~\ref{table:Main} and \ref{table:main-visual-expanded}, MiniGPT-4 and Llava do not perform adequately on these tasks. These models are inherently at a disadvantage because they were designed for visual instruction tuning, which typically involves handling one visual instance at a time. Instructed visual assembly, in contrast, requires managing a large collection of at least 20 images per instruction, making these models unsuitable for the task. However, they were the closest available open-source models to serve as our baselines.

In the case of MiniGPT-4, when prompted with several images (our collection) and an instructed assembly task, it tends to produce generic captions for some images in the collection. We will showcase examples of these failure cases later in the manuscript. One notable advantage of MiniGPT-4 is the use of Q-former, which reduces the number of tokens needed to represent a single image. This allowed us to utilize this model without major modifications since our entire task fit within the context limit of MiniGPT-4. Conversely, for Llava-1.5~\cite{liu2023llava}, a single image is represented by 576 tokens; with a collection of 20 images, the input sequence increases to approximately 11,520 tokens, far exceeding the original design limitations of LLMs~\cite{openai2023gpt}. We attempted to use vanilla Llava-1.5 but encountered significant context limitations that prevented the model from producing any meaningful output. We then modified Llava to reduce the token count per image by trying methods such as averaging all tokens, max pooling, or using the [CLS] token. We determined that using the [CLS] token was most effective for our task. Despite these efforts, Llava-1.5 was still unable to perform assembly tasks adequately. Unlike MiniGPT-4, Llava-1.5 attempted to follow instructions and often produced valid output sequences, yet it struggled to generate correct results.

For GPT-4o, we developed a prompting strategy that involved overlaying the identifier ID on each image in the collection and displaying the images as a high-resolution grid of 5x4. This approach eliminated the need for overlaying images and text from Algorithm~\ref{algorithm:tokenizer} when prompting GPT-4o. This method was only feasible because GPT-4o can process high-resolution images; attempts with MiniGPT-4 and Llava-1.5 failed as the resolution was insufficient to display the entire collection with visible IDs in a mosaic format.

\paragraph{\textbf{Expanded results of Multilen Assembler.}} In Table~\ref{table:multilen-vista-expanded}, we showcase per-task results of GPT-4o and the Mutilen-\assembler{} from Figure~\ref{fig:length-analysis} in the main manuscript. We report the average across all lengths. As observed in Figure 5 in the main manuscript, the Multilen-\assembler{} outperforms GPT-4o on all assembly tasks. 
\begin{table}[ht]
\centering
\footnotesize
\setlength{\tabcolsep}{2pt}
\renewcommand{\arraystretch}{1.2}
\caption{\textbf{Multilen-\vista{}}. Extended results of Figure~5 in main paper. We show the results per task of GPT-4o and the Multilen-\assembler{} on the Multilen-\vista{} dataset.}
\label{table:multilen-vista-expanded}
\begin{tabular}{lcccc|cccc|c}
        \cline{2-10}
         \multicolumn{1}{c}{} & \multicolumn{9}{c}{\textbf{Assembly Accuracy (\%)}} \\
        & \multicolumn{4}{c|}{\textbf{Positional Cues}} & \multicolumn{4}{c|}{\textbf{Semantic Cues}} &  \\
         & Ins. & Rem. & Repl. & Swap & Ins. & Rem. & Repl. & Swap & \textbf{Avg.} \\
         \hline
        GPT-4o & 35.0&	77.5&	31.3&	62.5&	8.8&	13.8&	12.5&	8.8& 31.3 \\
        \assembler{}-7B &	\textbf{98.8}&	\textbf{96.3}&	\textbf{98.8}&	\textbf{98.8}&	\textbf{51.3}&\textbf{	53.8}&	\textbf{48.8}&	\textbf{75.0}& \textbf{77.7} \\
        \hline
\end{tabular}

\end{table}

\paragraph{\textbf{Compositional \assembler{} performance at single tasks.}} In Table~\ref{table:compositional}, we showcase the performance of the Compositional Assembler on single tasks. We show results only on semantic cues since the model was trained on compositions of semantic tasks. In table~\ref{table:compositional-expanded} we show the results on all tasks. Interestingly, the results show that adding compositional tasks during training improves the average performance on \vst{}. Note that the Compositional \assembler{} was also trained on variable input timeline lengths, so it also performance relatilevely well on Multilen-\vista{}. 
\begin{table}[ht]
\centering
\setlength{\tabcolsep}{1pt}
\footnotesize
\renewcommand{\arraystretch}{1.2}
\caption{\textbf{Compositional \assembler{} results}. We showcase the results of the Compositional \assembler{} on semantic tasks on \vista{} and Multi-len-\vista{}.}
\label{table:compositional-expanded}
\begin{tabular}{lcccc|c}
        \cline{2-6}
         \multicolumn{1}{c}{} & \multicolumn{5}{c}{\textbf{Assembly Accuracy (\%)}} \\
         &  \multicolumn{4}{c}{\textbf{Semantic Cues}} &  \\
         & Insertion & Removal & Replacement & Swap & \textbf{Average} \\
         \hline
         \rowcolor[gray]{0.9}\multicolumn{6}{l}{\textbf{\vista{}}} \\
         
        \assembler{} & \textbf{42.5} & 55.0 & 41.7 & 27.9 & 41.8 \\
        Compositional \assembler{}  & 40.00& \textbf{58.75}&	\textbf{42.50}&	\textbf{35.00}&	\textbf{44.06} \\
        \hline
        \rowcolor[gray]{0.9}\multicolumn{6}{l}{\textbf{Mutlilen-\vista{}}} \\
        
        Multilen \assembler{} &\textbf{51.3}&	\textbf{53.8}&	\textbf{48.8}&	\textbf{75.0}& \textbf{57.2} \\
        Compositional \assembler{} & 48.8&	51.2&	45.0&	40.0&	46.2 \\
        \hline
\end{tabular}
\end{table}

\begin{table*}[ht]
    \centering
    \setlength{\tabcolsep}{3pt}
    \renewcommand{\arraystretch}{1.2}
    \footnotesize
    \caption{\textbf{Impact of Collection Size on the \assembler{} trained with Collection Size 20.} We show the impact of increasing the Collection Size of a model trained with a Collection Size equal to 20 (the original model is shown as gray in the row). We evaluate the model by increasing the Collection Size without training it. We observe the performance degrades when the Collection is more than double what the model was trained for. }
    \label{table:collection-size-expanded}
    \begin{subtable}{\linewidth}
        \centering
        \begin{tabular}{c|cccc|cccc|c}
        & \multicolumn{4}{c}{\textbf{Positional Cues}} & \multicolumn{4}{c}{\textbf{Semantic Cues}} &  \\
        Collection Size & Ins. & Rem. & Repl. & Swap & Ins. & Rem. & Repl. & Swap & \textbf{Avg.} \\
        \hline
        \textcolor{gray}{20} &	\textcolor{gray}{98.8}&	\textcolor{gray}{90.4}&	\textcolor{gray}{85.4}&	\textcolor{gray}{85.8}&	\textcolor{gray}{65.8}&	\textcolor{gray}{66.7}&	\textcolor{gray}{54.2}&	\textcolor{gray}{45.8}&    \textcolor{gray}{74.1}\\
        \hline
        30 &	97.5&	89.6&	84.6&	86.3&	56.3&	62.1&	43.8&	39.2&	69.9\\
        40 &	86.3&	86.7&	85.0&	85.4&	36.7&	52.9&	34.2&	25.4&	61.6\\
        50 &	50.0&	66.3&	77.9&	59.6&	19.2&	38.3&	25.8&	14.2&	43.9\\
        60 &	4.2&	29.6&	0.0&	17.5&	0.0&	6.7&	0.0&	0.0&	7.2\\
        \hline
    \end{tabular}
        \caption{Impact of Collection Size on \vst{}}
        \label{subtable:collection-size-vst-expanded}
    \end{subtable}%
    \\
    \begin{subtable}{\linewidth}
        \centering
        \begin{tabular}{c|cccc|cccc|c}
            & \multicolumn{4}{c}{\textbf{Positional Cues}} & \multicolumn{4}{c}{\textbf{Semantic Cues}} &  \\
            Collection Size & Ins. & Rem. & Repl. & Swap & Ins. & Rem. & Repl. & Swap & \textbf{Avg.} \\
            \hline
            \textcolor{gray}{20} & \textcolor{gray}{75.4} & \textcolor{gray}{98.8} & \textcolor{gray}{95.8} & \textcolor{gray}{97.1} & \textcolor{gray}{42.5} & \textcolor{gray}{55.0} & \textcolor{gray}{41.7} & \textcolor{gray}{27.9} & \textcolor{gray}{66.8} \\

            \hline
            30 & 80.4 & 97.1 & 93.3 & 95.8 & 28.3 & 49.2 & 45.4 & 21.3 & 63.9 \\
            40 & 72.9 & 90.8 & 93.8 & 88.8 & 27.5 & 40.0 & 31.7 & 20.0 & 58.2 \\
            50 & 40.4 & 50.0 & 64.2 & 53.3 & 13.8 & 26.7 & 17.9 & 14.2 & 35.1 \\
            60 & 12.9 & 33.3 & 32.5 & 39.6 & 0.0 & 9.6 & 0.0 & 1.3 & 16.2 \\
            \hline
        \end{tabular}
        \caption{Impact of Collection Size on \vista{}}
        \vspace{-20pt}
        \label{subtable:collection-size-vista-expanded}
    \end{subtable}
\end{table*}

\paragraph{\textbf{Performance at different collection sizes.}}
In Table~\ref{table:collection-size-expanded}, we demonstrate the \assembler{}-7B model's performance, which was trained with a collection size (\( |C| \)) of 20, when tested with varying \( |C| \). As expected, a gradual decrease in performance was observed as \( |C| \) increases. Notably, up to \( |C|=40 \), \assembler{} maintains reasonable performance despite not being trained for larger \( |C| \). However, at \( |C|=50 \) and more significantly at \( |C|=60 \), we see a substantial drop in performance, making the model less effective for the task. This finding could be due to the increased difficulty in locating a specific item as \( |C| \) grows, since the chance of randomly finding an item is \( 1/|C| \). Additionally, the original Llama model's~\cite{touvron2023llama} context size limit may contribute to this effect. When \( |C| \) exceeds 40, the used context approaches 2048 tokens, the Llama's limit. To mitigate this issue, we employ RoPE scaling~\cite{kaiokendev2023extending} to double the context to 4096 tokens. Despite RoPE scaling's relative effectiveness, it does incur some performance degradation as \( |C| \) increases, as explained in~\cite{kaiokendev2023extending}. Therefore, we suggest the dramatic performance drop at \( |C|=60 \) is due to both the increased task complexity and the impact of context scaling. We believe future research in enabling longer context in LLMs will empower our Timeline Assembler to handle much larger collection sizes.

\paragraph{\textbf{Comparison with Transcript to Video\cite{xiong2022transcript}.}} We classify assembly instructions using a GPT-4o ``task classifier'' as described in our Section \ref{subsec:capabilities} (Table \ref{table:multitask}). Second, we extract BLIP captions for each visual asset in the timeline. We then use CLIP text embeddings to find and edit the prompt according to the assembly instruction. Our re-implementation of TtV, builds on the stronger CLIP backbone (instead of S3D). When compared ins VIST-A, we observe that TtV underperforms compared to our model (39.7\% vs. 81.6\%), likely due to compounding errors introduced by each step of the system.

\paragraph{\textbf{MiniGPT-4 Failure Mode.}} We extended MiniGPT-4 to our task by enabling it to accommodate multiple visual elements as input, as described in Section \ref{sec:method}. Despite this adaptation, MiniGPT-4 was unable to comprehend assembly instructions, resulting in 0\% assembly accuracy and often producing text related to isolated assets in the collection without forming any interpretable output timeline (please see Figure \ref{fig:minigpt4answers}). These results emphasize the novelty/contribution of our work, which lies in: (i) adapting multimodal LLMs to handle sets (the input collection) and sequences (the input and output timelines) of visual assets (video clips or images), and in (ii) developing a unique instruction fine-tuning pipeline that uses automatically generated assembly instructions for training the visual assembly task.

\begin{figure}
\centering
\begin{subfigure}{0.85\linewidth}
\begin{tcolorbox}[colback=gray!5!white,colframe=gray!75!black,title=Figure~\ref{fig:qualitative_result_004} example output.]

\textbf{Instruction:} Remove the ``second"' clip and replace it with the clip ``showing a woman pouring water into a bowl of berries"".
\begin{lstlisting}
{ 1: {"clip_id": "4864"},2: {"clip_id": "6085"}, 
3: {"clip_id": "3651"}, 4: {"clip_id": "7257"}, 
5: {"clip_id": "7580"}, 6: {"clip_id": "3515"} }
\end{lstlisting}

\textbf{MiniGPT4 Output:} `“a mushroom”'

\end{tcolorbox}
\caption{\textbf{Output for sample in Fig~\ref{fig:qualitative_result_004}.} MiniGPT4 simply generates the phrase `a mushroom', showing its inability to perform assembly tasks out of the shelf.}\label{figure:example-input}
\end{subfigure}

\begin{subfigure}{0.85\linewidth}
\begin{tcolorbox}[colback=gray!5!white,colframe=gray!75!black,title=Figure~\ref{fig:qualitative_result_001} example output.]

\textbf{Instruction:} Add the clip of the "bird on a tree" in the  the "first" position of the timeline. Then, remove the clip of the "dog rolling in the grass".
\begin{lstlisting}
{1: {'clip_id': '0975'},2: {'clip_id': '7182'}, 
 3: {'clip_id': '6278'},4: {'clip_id': '1817'},
 5: {'clip_id': '9096'}}
\end{lstlisting}

\textbf{MiniGPT4 Output:}  `The dog's on a tree is a "92" in the position of the timeline. Add the clip of the "dog crawling on grass" in the "first" position of the timeline, and remove the clip of the "dog crawling on grass" in the "second" position.'

\end{tcolorbox}
\caption{\textbf{Output for sample in Fig~\ref{fig:qualitative_result_001}.} In this case, MiniGPT4 gives some descriptions related to the iamge and the instruction, but does not output the updated timeline.}\label{figure:example-input}
\end{subfigure}




\caption{\textbf{Examples of MiniGPT-4 Outputs.} We showcase three examples of MiniGPT4 outputs for the assembly samples in our dataset. MiniGPT4 yields close to 0\% assembly accuracy due to its inability to understand the assembly tasks. To observe how the \assembler{} succesfully performs the assembly tasks please refer to figures in Section~\ref{sec:qual-supp}.}
\label{fig:minigpt4answers}
\end{figure}
\clearpage
\section{Input Tokenizer}
\label{sec:tokenizers}

\paragraph{Visual-Text Inputs.} In Section 3 of the main paper, we explained how we manage the visual elements through a collection that contains all the visual information. This modeling allows us to manage the timeline as pointers to the visual elements. To clarify how exactly we do this process, please refer to Figure~\ref{figure:example-input} that shows an example before being tokenized and fed to the LLM. Figure~\ref{figure:example-input} shows the basic structure of the collection containing the fields  \texttt{clip\_id} and \texttt{clip}. We use these two fields to store the information of each clip, such that \texttt{clip\_id} contains the identifier token $\mathbf{x}^i_k$ for every clip, and \texttt{clip} contains the visual information $\mathbf{x}^i_v)$ which stores the visual tokens coming from the output of the projection layer $h_{\gamma}$. Algorithm~\ref{algorithm:tokenizer} shows how we parse each input to place the visual token in their corresponding fields in the collection.
In a nutshell, Algorithm~\ref{algorithm:tokenizer} splits the inputs into several pieces based on the string ``\textless\texttt{VisualHere}\textgreater''. For each piece we know that there is a visual token separating them (given the structure of the collection shown in Figure~\ref{figure:example-input}). Thus, we simply find all the ``\textless\texttt{VisualHere}\textgreater'' placeholders and replace them with the visual tokens coming the visual backbone $g$ and the projection layer $h_{\gamma}$. The rest of the input is simply processed as standard text.
\begin{algorithm}
\caption{Visual-Text Tokenizer Function} \label{algorithm:tokenizer}
\begin{algorithmic}
\Require $collection$, $prompt$
\Ensure $prompt$ is a string
\Ensure $collection$ is a list
\Ensure $collection[i]$ is an Image

\State $prompt\_splits \gets \text{split}(prompt, \text{"\textless VisualHere\textgreater"})$
\State $sequence \gets \text{empty list}$

\For{$i \gets 0$ \textbf{to} $\text{length}(prompt\_splits) - 1$}
    \State $prompt\_piece = prompt\_splits[i]$
    \State $prompt\_piece\_emb = \text{llm.embed}(prompt\_piece)$
    
    \State $\text{append}(sequence, prompt\_piece\_emb)$
    
    \If{$i < \text{length}(prompt\_splits) - 1$}
        \State $visual\_embed = \text{vit.embed}(collection[i])$
        \State $\text{append}(sequence, visual\_embed)$
    \EndIf
\EndFor
\State \Return $sequence$
\end{algorithmic}
\end{algorithm}
\vspace{-15pt}

\begin{figure}
\begin{tcolorbox}[colback=gray!5!white,colframe=gray!75!black,title=Input Example]
\textbf{Human:} You are presented with a collection of images, you have to modify the timeline accordingly, to complete the instruction you are given.

\textbf{Collection:}
\begin{lstlisting}
[
{"clip_id":0935,"clip":"<VisualHere>"},
{"clip_id":0721,"clip":"<VisualHere>"},
...
{"clip_id":0612,"clip":"<VisualHere>"},
{"clip_id":0899,"clip":"<VisualHere>"},
...
{"clip_id":0327,"clip":"<VisualHere>"}
]
\end{lstlisting}

\textbf{The current timeline is:}
\begin{lstlisting}
{
    1: {"clip_id": 0894},
    2: {"clip_id": 0374},
    3: {"clip_id": 0899},
    4: {"clip_id": 0464},
    5: {"clip_id": 0665}
}
\end{lstlisting}

\textbf{Instruction:} Remove the ``second'' clip and replace it with the clip ``showing a woman pouring water into a bowl of berries''.

\begin{lstlisting}
{
    1: {"clip_id": 0894},
    2: {"clip_id": 0374},
    3: {"clip_id": 0899},
    4: {"clip_id": 0464},
    5: {"clip_id": 0665}
}
\end{lstlisting}

\end{tcolorbox}
\caption{\textbf{Input Example.} We show an example of the whole input required to perform assembly tasks. This input includes the Collection, the Input Timeline, and the Instruction that tells the model what kind of operation to perform on the Timeline.}
\vspace{-5pt}
\end{figure}
\paragraph{Text-Only Inputs.} For text-only baselines, we replace the string ``\textless\texttt{VisualHere}\textgreater'' with the BLIP-2~\cite{li2023blip} generated caption of the corresponding visual asset, and tokenize the entire input as text. In this case, Algorithm~\ref{algorithm:tokenizer} is not needed.

\section{Implementation Details}
\label{sec:implementation}
\paragraph{Additional Training Details.} 
In Table~\ref{tab:hyperparameters} we show more training details, including the LoRA~\cite{hu2021lora} hyper-parameters. Our model trains on 1 GPU A100 for 100k iterations, which is around ~16hours of compute per training. At test time, the model takes around 5 seconds per instruction on the same hardware. 
\paragraph{Assembly Task Classifier with GPT4o.} 
In the multi-task analysis (Table~\ref{table:multitask}) of the main paper we used a task-classifier based on GPT-4o to decide which model to use depending on the input query. We use the input prompt shown in Figure~\ref{figure:gpt-3.5-classifier} to query GPT-4o and classify into one of eight tasks, each of the assembly instructions in our testing set. The results of this classifier are shown in Table~\ref{tab:gpt-3.5-classifier}.  We observed that the Semantic Swap task can be challenging to identify, which might be one of the reasons why this task is the lowest-performing one across all models.
\noindent\begin{minipage}{.45\linewidth}
\centering
\footnotesize
\setlength{\tabcolsep}{0.5pt}
\renewcommand{\arraystretch}{1}
\captionof{table}{\textbf{Training Parameters}}
\label{tab:hyperparameters}
\begin{tabular}{@{}lc@{}}
\toprule
Configuration         & Value \\ \midrule
LoRA - r              & 16 \\
LoRA - $\alpha$         & 32 \\
LoRA - dropout        & 0.1 \\
\midrule
Optimizer             & AdamW \\
Optimizer momentum    & \(\beta_1, \beta_2 = 0.9, 0.999\) \\
Weight decay          & 0.05 \\
Base learning rate    & \(6 \times 10^{-5}\) \\
Warmup learning rate &  \(5 \times 10^{-7}\) \\
Minimum learning rate & \(2 \times 10^{-6}\) \\
Learning rate schedule & cosine decay \\
Warm-up iterations     & \textit{VST},\textit{VISTA}$=$$7K$,$10K$\\
Total iterations      & \textit{VST},\textit{VISTA}$=$$70K$,$100K$\\
Batch size            & 1 \\ \bottomrule
\end{tabular}
\end{minipage}%
\hfill
\begin{minipage}{.4\linewidth}
\centering
\footnotesize
\setlength{\tabcolsep}{5pt}
\renewcommand{\arraystretch}{1.2}
\captionof{table}{\textbf{GPT-4o Task-Classifier.}}
\label{tab:gpt-3.5-classifier}
\begin{tabular}{l|c}
\hline
\toprule
\textbf{Task}           & \textbf{Accuracy} \\ \hline
Positional Insertion  & 88.8\%      \\
Positional Removal    & 98.8\%     \\
Positional Replacement & 80.0\%     \\
Positional Swap       & 77.5\%      \\
Semantic Insertion    & 95.0\%      \\
Semantic Replacement  & 66.2\%      \\
Semantic Removal      & 100.0\%      \\
Semantic Swap         & 46.2\%      \\
\bottomrule
\end{tabular}

\end{minipage}

\begin{figure}
\begin{tcolorbox}[colback=blue!5!white,colframe=blue!75!black,title=GPT-4o Classifier: Input Prompt]
\scriptsize{
\textbf{TEMPLATE} = You task is to classify an assembly instruction into 8 different categories. 
There are four different types of operations:

\textbf{Insert:} Inserts an element from the visual collection into the timeline.

\textbf{Remove:} Deletes an element from the timeline.

\textbf{Replace:} Changes an element from the timeline with one from the collection.

\textbf{Swap:} Interchanges two elements from the timeline. 

We can refer to the visual elements by: 

\textbf{(a) a semantic reference}, which describes the visual appearance of the elements with text, and 

\textbf{(b) ID reference}, which specifies the elements to manipulate by their unique ID and position in the timeline. 

That said, in total there could be 8 categories, which we provide an example for each below: 

\textbf{positional\_insertion:} ``Append the shot with ID 34 to the end of the sequence''

\textbf{positional\_removal:} ``Remove the last shot''

\textbf{positional\_replacement:} ``Replace the shot 16 with shot 58''

\textbf{positional\_swap:} ``Exchange the second and third shot''

\textbf{semantic\_insertion:} ``Add the shot where the girl is eating ice-cream to the end of the sequence''

\textbf{semantic\_removal:} ``Delete the shot of the girl in her room''

\textbf{semantic\_replacement:} ``Change the shot of the girl entering the car with the one of the girl and the goat''

\textbf{semantic\_swap:} ``Swap the shot of the girl eating ice-cream with the last shot of the sequence''

IMPORTANT: Just output the query class. DO NOT explain.

\textbf{Query:} ``\{ \}''
}
\end{tcolorbox}
\caption{\textbf{Input query for GPT-4o Task Classifier.}}\label{figure:gpt-3.5-classifier}
\end{figure}

\clearpage
\newpage
\section{Failure Modes}\label{sec:failure}

We include failure modes of the \assembler{} in Figures~\ref{fig:qualitative_result_001}~to~\ref{fig:qualitative_result_004}. We explain each failure mode in details inside the red boxes. In general, the \assembler{}'s failures often occur due to misinterpretation of instructions (Figures~\ref{fig:qualitative_result_002},\ref{fig:qualitative_result_003},\ref{fig:qualitative_result_004}), not finding elements in the collection that match the visual description (Figures~\ref{fig:qualitative_result_001},\ref{fig:qualitative_result_003},\ref{fig:qualitative_result_004}), or executing only one of the of the two tasks it was instructed to do (Figures~\ref{fig:qualitative_result_002}, and \ref{fig:qualitative_result_003}).

\section{Additional Qualitative Results.}\label{sec:qual-supp}

\begin{figure*}[h!]
    \centering
    \includegraphics[width=0.98\linewidth]{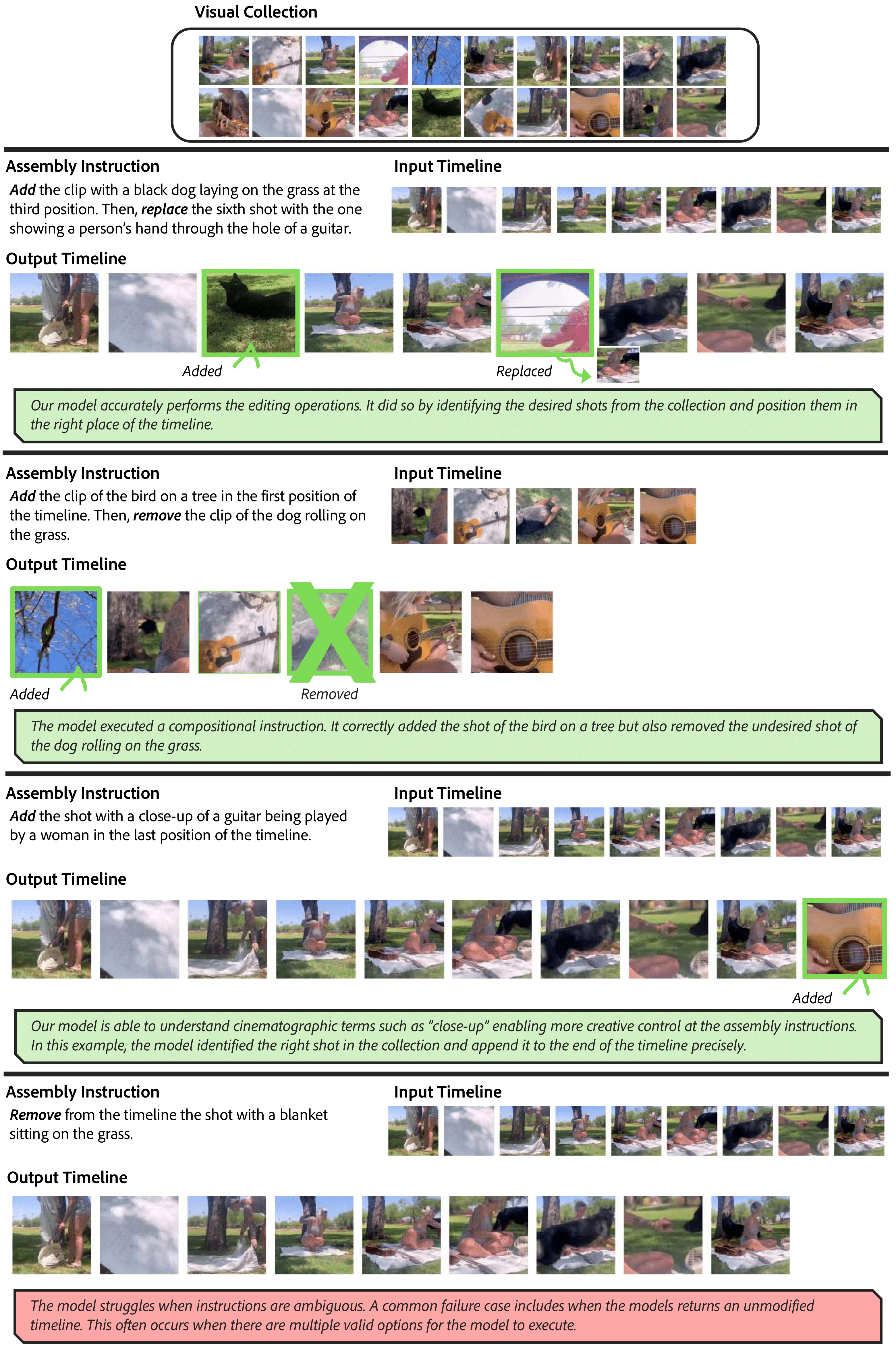}

    \caption{\textbf{Qualitative Results 1.} We showcase an extended version of the first example in the main paper. We showcase its collection and additional examples.}
    \label{fig:qualitative_result_001}
\end{figure*}

\begin{figure*}[h!]
    \centering
    \includegraphics[width=0.98\linewidth]{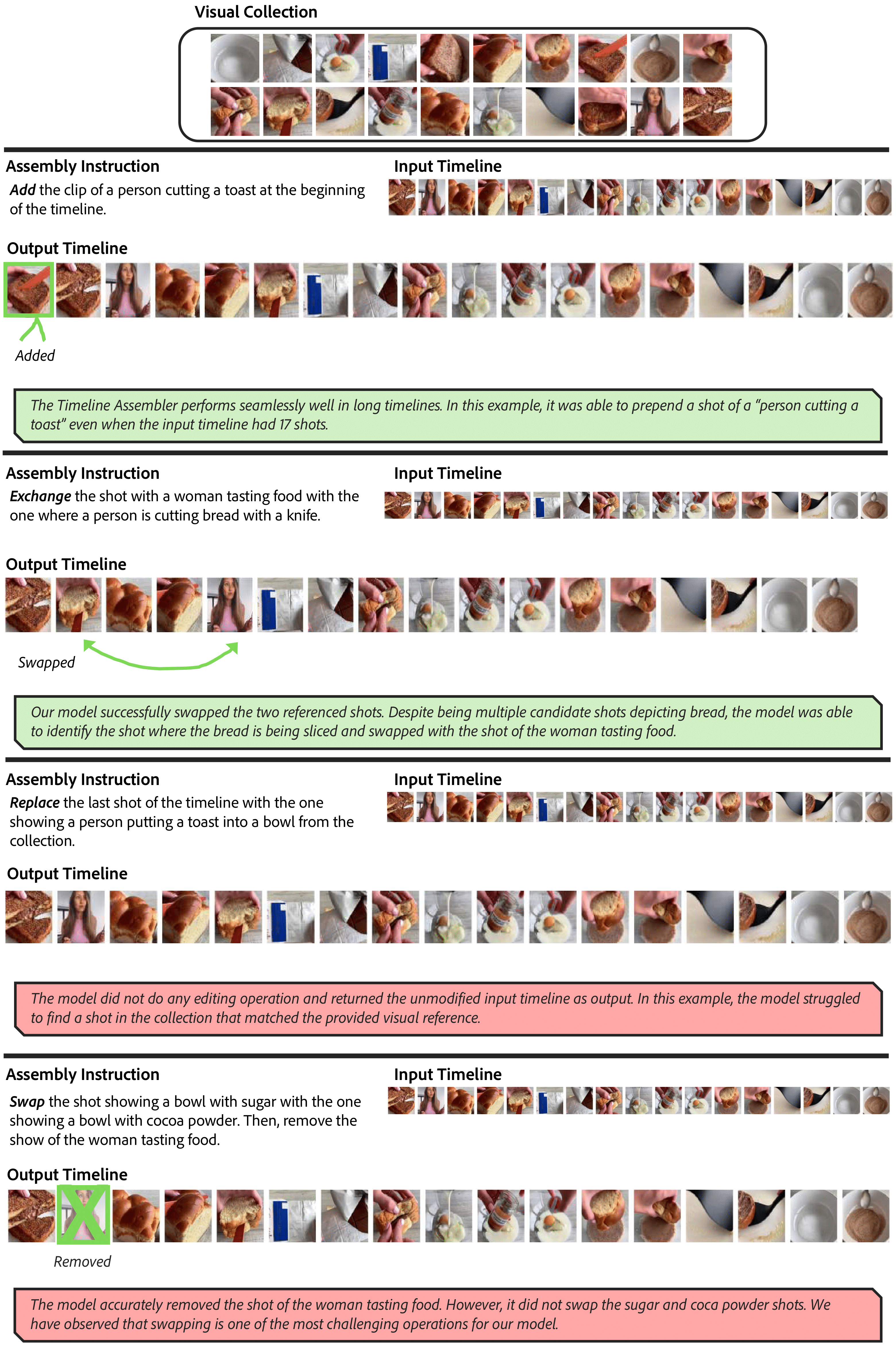}

    \caption{\textbf{Qualitative Results 2.} We showcase an extended version of the second example in the main paper. We showcase its collection and additional examples.}
    \label{fig:qualitative_result_002}
\end{figure*}

\begin{figure*}[h!]
    \centering
    \includegraphics[width=0.98\linewidth]{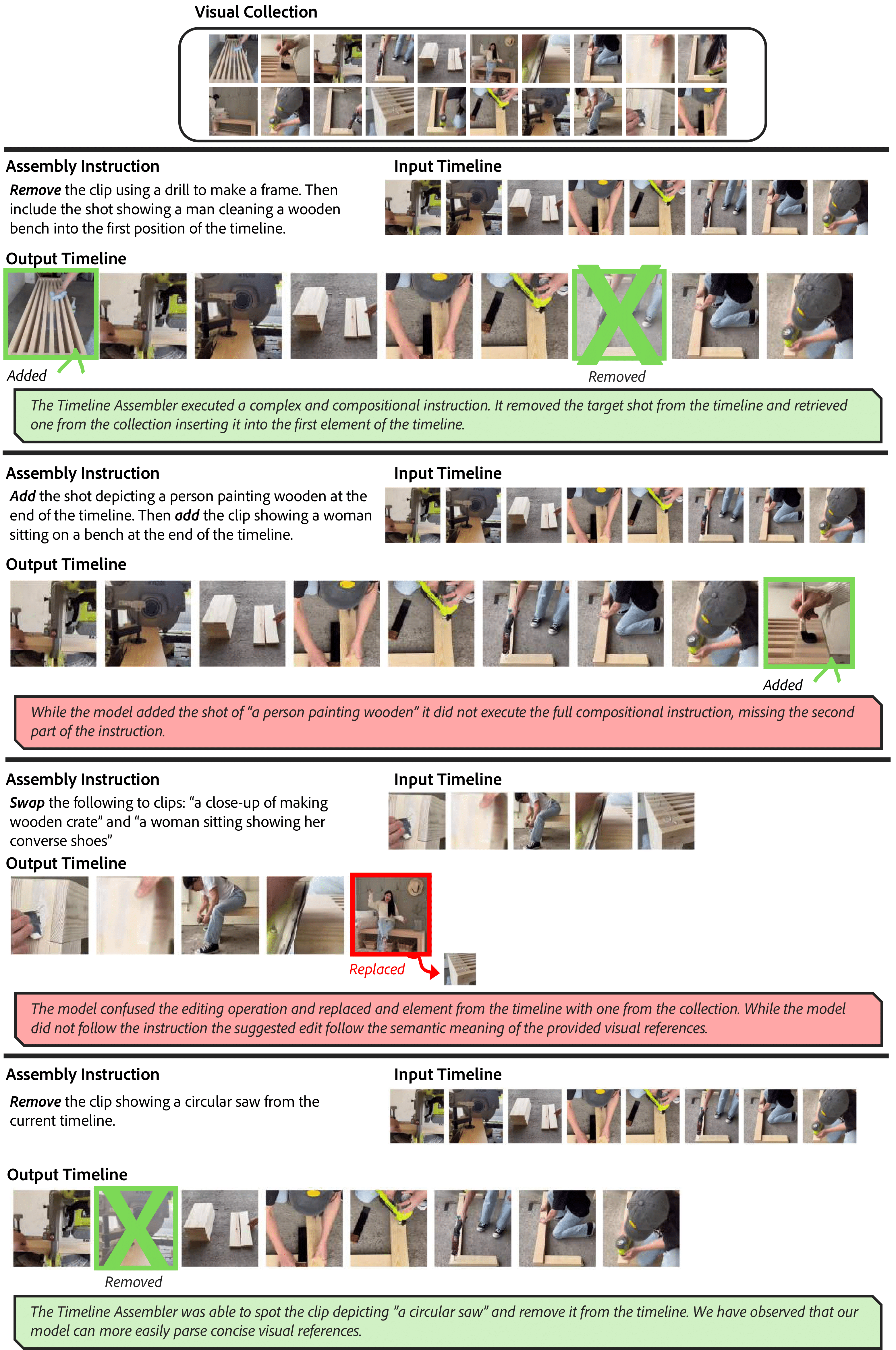}

    \caption{\textbf{Qualitative Results 3.} We showcase an extended version of the third example in the main paper. We showcase its collection and additional examples.}
    \label{fig:qualitative_result_003}
\end{figure*}

\begin{figure*}[h!]
    \centering
    \includegraphics[width=0.98\linewidth]{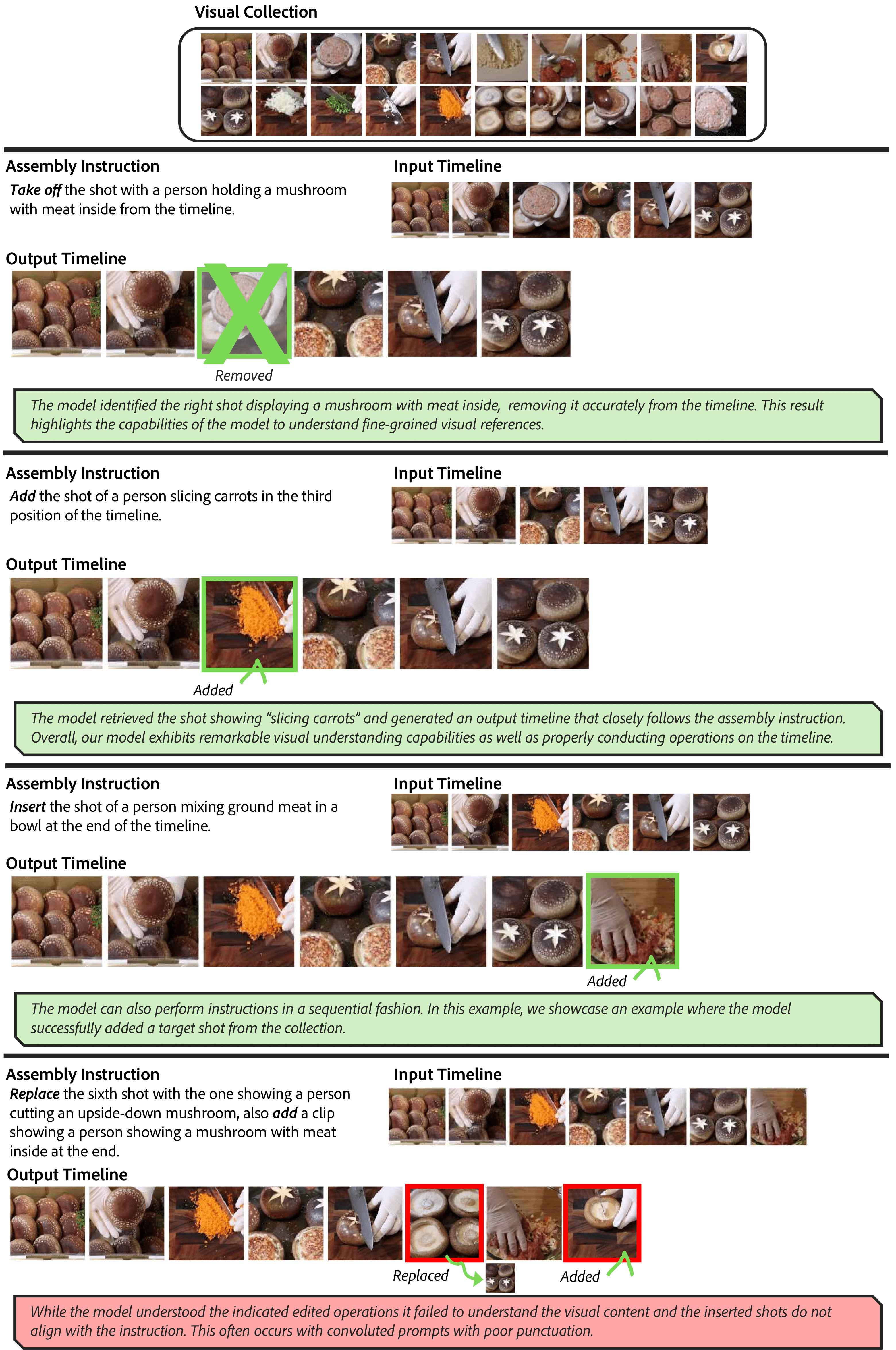}

    \caption{\textbf{Qualitative Results 4.} We showcase an extended version of the fourth example in the main paper. We showcase its collection and additional examples.}
    \label{fig:qualitative_result_004}
\end{figure*}

In Figures~\ref{fig:qualitative_result_001}~to~\ref{fig:qualitative_result_004} we show more qualitative results. We include the collections that we omitted in the main paper due to space reasons. We can see the Collections along with some examples in the same order that they appeared in the main paper. We show 3 additional examples that showcase the \assembler{}'s ability to manage timelines and modify them given a wide variety of instructions.


\end{document}